\documentclass{article}

\PassOptionsToPackage{numbers, compress}{natbib}

\usepackage[preprint]{neurips_2022}

\usepackage[utf8]{inputenc} 
\usepackage[T1]{fontenc}    
\usepackage{hyperref}       
\usepackage{url}            
\usepackage{booktabs}       
\usepackage{amsfonts}       
\usepackage{nicefrac}       
\usepackage{microtype}      
\usepackage{xcolor}         

\usepackage{xspace}
\usepackage{graphicx}
\usepackage{amsmath,amssymb} 
\usepackage{color}
\usepackage{subfig} 
\usepackage{multicol}
\usepackage{algorithm}
\usepackage{algorithmicx}

\newcommand*{\eg}{e.g.\@\xspace}
\newcommand*{\ie}{i.e.\@\xspace}
\newcommand*{\etc}{\emph{etc}.\@\xspace}

\newcommand*{\etal}{\emph{et al}.\@\xspace}

\newcommand{\ruoteng}[1]{\textcolor[rgb]{0.3,0.8,0.1}{#1}}

\title{Realistic Large-Scale Fine-Depth Dehazing Dataset from 3D Videos}

\author{%
  Ruoteng Li \\
  ByteDance \& \\ National University of Singapore \\  
  \And
  Xiaoyi Zhang \\
  National University of Singapore \\  
  \And
  Shaodi You \\
  University of Amsterdam \\  
  \And
  Yu Li \\
  IDEA \& UIUC \\  
}

\begin{document}

\maketitle

\begin{abstract}
Image dehazing is one of the important and popular topics in computer vision and machine learning. A reliable real-time dehazing method with reliable performance is highly desired for many applications such as autonomous driving, security surveillance, \etc While recent learning-based methods require datasets containing pairs of hazy images and clean ground truth, it is impossible to capture them in real scenes.  Many existing works compromise this difficulty to generate hazy images by rendering the haze from depth on common RGBD datasets using the haze imaging model. However, there is still a gap between the synthetic datasets and real hazy images as large datasets with high-quality depth are mostly indoor and depth maps for outdoor are imprecise. In this paper, we complement the existing datasets with a new, large, and diverse dehazing dataset containing real outdoor scenes from High-Definition (HD) 3D movies. We select a large number of high-quality frames of real outdoor scenes and render haze on them using depth from stereo. Our dataset is clearly more realistic and more diversified with better visual quality than existing ones. More importantly, we demonstrate that using this dataset greatly improves the dehazing performance on real scenes. In addition to the dataset, we also evaluate a series state of the art methods on the proposed benchmarking datasets. \footnote{The rendered haze dataset will be available upon request.}
\end{abstract}

\section{Introduction}
\label{sec:intro}

Haze is one of the most common bad weather phenomena caused by floating atmospheric particles that degrade the contrast and visibility of images captured outdoor.  Many vision-based algorithms deployed in outdoor environments suffer from hazy conditions such as object detection \cite{object_detection_haze}, semantic segmentation \cite{SDV18}, visual tracking \cite{visual_tracking_haze_2014,target_tracking_hazy_2018}, and so on.  It is important to enhance the visibility of hazy images. 

Since \cite{tan2008} with the first single image dehazing algorithm proposed , a large number of haze removal methods have been proposed \cite{fattal2014,li2019reside,tarel2012frida}. While traditional model-based dehazing algorithms, in recent years, data driven methods require the training datasets with high quality ground truth and their performance heavily rely on the quality of the training data. To that end, a number of dehazing datasets are proposed, most of which are synthetic datasets rendered based on the optical model for particle scattering known as Koschmieder's law \cite{koschmieder1925theorie}. The light luminance $\mathbf{I}$ captured at a certain pixel $\mathbf{x} = (x, y)$ can be expressed as:
\begin{equation}
\mathbf{I}(\mathbf{x}) = \mathbf{J}(\mathbf{x})
e^{-\beta d(\mathbf{x})} + \mathbf{A} (1-e^{-\beta d(\mathbf{x})}), 
\label{eq:scattering_model}
\end{equation}
where $\mathbf{J}$ represents the light emitted from the scene object, namely the haze free image, which is attenuated by the scattering media with a coefficient $e^{-\beta d(x)}$. $d(\mathbf{x})$ represents the absolute depth value at location $\mathbf{x}$ and the scattering coefficient of the medium is defined as $\beta$. $\mathbf{A}$ is the atmospheric light. 

As can be seen, the rendering pipeline requires accurate depth $d(\mathbf{x})$. However, it is usually very difficult to obtain accurate and dense depth maps from outdoor scenes. Some of the existing datasets collect RGBD images with the aid of infrared \cite{Lthen2017ARD}, but infrared is usually lack of accuracy in wild outdoor scenes. Other datasets \cite{li2019reside} apply the latest depth estimation algorithms. However, depth estimation algorithms do not provide high quality scene depth because depth estimation itself is still an open research topic. 

\begin{figure*}[t!]
\centering
 \vspace{-10pt}
\includegraphics[width=1.0\linewidth]{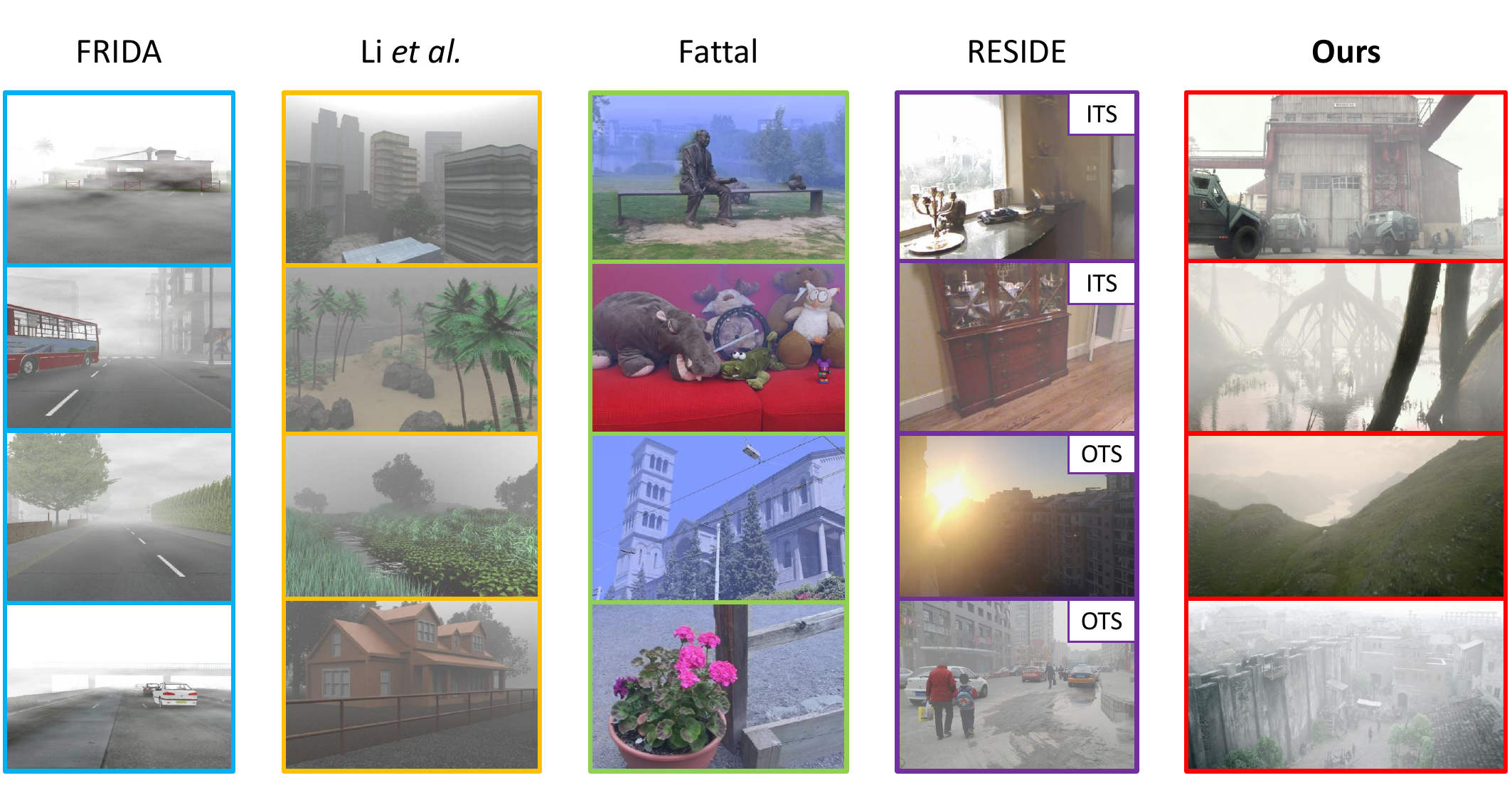}
\vspace{-10pt}
\caption{Visual comparison of the existing datasets including Frida \cite{tarel2012frida}, Li~\etal~\cite{li2017survey}, Fattal \cite{fattal2008}, RESIDE \cite{li2019reside}, and our proposed dataset \textbf{LSFD}. ITS and OTS indicate indoor training set and outdoor training set respectively.}
\label{fig:example_compare_datasets}
 \vspace{-15pt}
\end{figure*}

In this paper, we argue that a large-scale high-quality and realistic dataset is the bottleneck for current dehazing algorithm. Especially, accurate and physically-valid ground truth is crucial for effectively training a dehazing method. 

First, most datasets are too small for modern data driven methods.  Some of the datasets (\eg, ~\cite{fattal2014}, ~\cite{tarel2012frida} and ~\cite{li2017survey}) contain only dozens of samples, which are not enough for learning-based methods. Second, the quality of the depth of existing outdoor datasets is undesirable. For example, \cite{li2019reside}  is a large-scale indoor and outdoor dehazing dataset. The synthetic haze is rendered on indoor scenes, which differs from the real-world scenarios. In addition, the outdoor subset of~\cite{li2019reside} applies monocular depth estimation to obtain the depth. However, the estimated depth is neither physically valid nor accurate, causing the rendered hazy images to appear as unrealistic.  Lastly, the existing outdoor datasets are unrealistic. As proved by our benchmark, training on unrealistic dataset limits the generalizability of data-driven models drastically.

We tackle the aforementioned problems using High-Definition (HD) videos. Based on the physics modeling in Eq.~\ref{eq:scattering_model}, we propose a large, high quality, highly diversified dataset. To create this dataset, we select 2000 high quality ground-truth images from 3D movies with 1920 $\times$ 1080 resolution. The images are captured by high-end cameras where the color, exposure, and sensor noise are optimized. The 2000 background images are from 22 different movies with 40 hours in total to ensure their diversity. More importantly, all those movies are captured by multi-view stereo cameras, so that we can obtain high quality depth using multi-view stereo enables us to render the high quality and physics realistic hazy images. Fig.~\ref{fig:example_compare_datasets} shows some examples and comparisons with existing datasets. Details of the dataset creation are described in Sec.~\ref{sec:dataset}.

Based on it, this paper proposes a systematic and detailed benchmark using the proposed dataset and existing datasets. We believe this work will enable more exciting and novel topics in dehazing.

\begin{table*}[!t]
    \centering
    \vspace{-5pt}
    \footnotesize
    \setlength{\tabcolsep}{3pt}
    \caption{Comparison of the proposed dataset with existing dehazing datasets. Our dataset contains the most diverse and accurate depth. }
    \begin{tabular}{lccccccc}
    \toprule
    {}    & Indoor    & Outdoor   & Background    & Diversity  & Haze effect   & Depth accuracy & Resolution \\
    \midrule
    Fattal~\cite{fattal2014}    & 4 & 8     & Real  & Low    & Using depth   & High  & 624 $\times$ 416\\
    FRIDA~\cite{tarel2012frida} & - & 480   & Synthetic & Low & Using depth & High   & 640 $\times$ 480 \\
    Li~\etal~\cite{li2017survey}& -  & 5     & Synthetic & Low & Ray tracing &   -   & 640 $\times$ 480 \\
    O-Haze~\cite{O_HAZE_2018}   & -  &  45   & Real      & High & Using depth & High & 3200 $\times$ 2900 \\
    RESIDE-ITS~\cite{li2019reside}& 1399 & - & Real      & Low & Using depth & High  & 620 $\times$ 460 \\
    RESIDE-OTS~\cite{li2019reside}& - &2061  & Real      & High & Using depth & Low  & 549 $\times$ 718 \\
    Foggy-CityScape~\cite{sakaridis2018semantic} & - & 5000 & Real & Low & Using depth & High & 1280 $\times$ 960 \\
    Foggy-Driving ~\cite{sakaridis2018semantic} & - & 101 & Real & Low & Using depth & High & 640 $\times$ 480 \\
    \hline
    \textbf{Ours}                 & - & 2000  & Real & High   & Using depth & High  & 1920 $\times$ 1080\\
    \bottomrule
    \end{tabular}
    \label{tab:datasets}
    \vspace{-10pt}
\end{table*}

The main contributions of this paper are summarized as:
\begin{itemize}
\item We create a large-scale outdoor dehazing dataset of real scenes.
\item The proposed dataset are developed based on high-definition (HD) images, rendered with physically-valid haze. 
\item We provide a systematical benchmark for state of the art dehazing methods on both  RESIDE \cite{li2019reside} dataset and the proposed dataset.
\end{itemize}

\section{Related Work}
\label{sec:related}
Haze removal from a single image has attracted wide attention. While early research works on dehazing using multiple images (\eg~\cite{zhuwen15,narasimhan2003}) or additional information from other sources (\eg~\cite{kopf2008,schechner2001instant}), modern methods work on single image using data-driven methods~\cite{Qu_2019_CVPR,ren2016mscnn,ren2018gfn,yang2018proximal,zhang2018dcpdn}. A comprehensive survey for dehazing can be found in~\cite{li2017survey}. In this work, we only focus on single image dehazing. 

\noindent \textbf{Dehazing datasets}
 Quantitative evaluation and supervised learning for dehazing require the hazy image and the pixel-wise corresponding clear image as a reference. These pairs are difficult to capture in real as it is usually impossible to control the outdoor environments. Alternatively, the most common approach is to render haze on clean images with its depth using Eq.~\ref{eq:scattering_model}~\cite{fattal2014,li2019reside,tarel2012frida}. An exception is the dataset of Li~\etal~\cite{li2017survey} that applies physics-based rendering (\ie~{ray tracing})~\cite{pharr2010physically} to generate the haze effect on synthetic scenes. Fattal~\cite{fattal2014} renders from real images and their depth maps. FRIDA~\cite{tarel2012frida} dataset focuses on driving scenarios and uses synthetic road backgrounds from graphics models. The datasets are small and are not suitable for learning-based methods. Recently, \cite{sakaridis2018semantic} introduces a foggy Cityscapes dataset using images from the Cityscapes dataset~\cite{cordts2016cityscapes}. However, the depth map quality in Cityscapes is poor in terms of considering is as haze transmission ground truth. Therefore the hazy images rendered from them are not satisfying. In addition, the Foggy-Driving dataset only contains 101 samples with varying image size. This dataset is only suitable for benchmarking dehazing methods.  \cite{O_HAZE_2018} introduces a small set of 45 fine-crafted hazy images, which are usually served well in evaluating dehazing methods rather than providing a training option for learning-based methods.  \cite{li2019reside} proposes a general dehazing dataset named RESIDE. Its indoor training set (ITS) is built on existing indoor RBGD datasets NYUv2~\cite{silberman2012nyu} and Middleburry stereo~\cite{scharstein2003high}. The outdoor training set (OTS) collects real outdoor images with their depth estimated using learning-based single image 2D to 3D estimation~\cite{liu2015learning}. The accuracy of learning-based depth is still not satisfying because it is much poorer than physically measured depths. However, the physically measuring is impractical currently due to the limitation of technical facilities. Other kinds of datasets like subjective evaluation datasets and task-driven datasets are not discussed here.

\noindent \textbf{Single image dehazing methods}
Early works on single image dehazing are mostly prior based methods and do not require training data~\cite{tan2008,meng2013bccr,tang2014,zhu2015cap,chen2016grm,li2014contrast,li2015nighttime,zhang2017fast,Sulami2014}. These works propose different priors in estimating the transmission map and the air light and then a clean image can be recovered by reversing the hazy image model in Eq.~\ref{eq:scattering_model}.  Among the works, \cite{he2011dcp} observes an interesting phenomenon of outdoor natural scenes with clear visibility and formulate as dark channel prior (DCP), which becomes one of the most successful priors to dehaze.  Unlike the previous local priors, \cite{berman2016nonlocal} develops a non-local prior which relies on the assumption that colors of a haze-free image form tight clusters in RGB space. Convolution neural network (CNN) based methods gain increasing popularity in recent years~\cite{galdran2018duality,cai2016dehazenet,li2017aod,li2018single,li2019lap,liu2019griddehazenet,liu2019ldp,Qu_2019_CVPR,ren2016mscnn,ren2018gfn,yang2018proximal,zhang2018dcpdn}. 
Methods~\cite{ren2018gfn,yang2018proximal} incorporate traditional dehazing priors and color processing operations into the deep network. Generative adversarial network~(GAN) based dehazing methods have been proposed in~\cite{li2018single,Qu_2019_CVPR}. The recent state-of-the-art method~\cite{liu2019griddehazenet,Wu_2021_contrast,Yuan_2020_domain,Peter_2019_Feature,Sourya_2020_Hierarchical,yu2021twobranch,chen2020pmhld} applies a novel attention-based multi-scale estimation on a grid network.


\begin{figure}[!t]
\centering
 \includegraphics[width=0.8\linewidth]{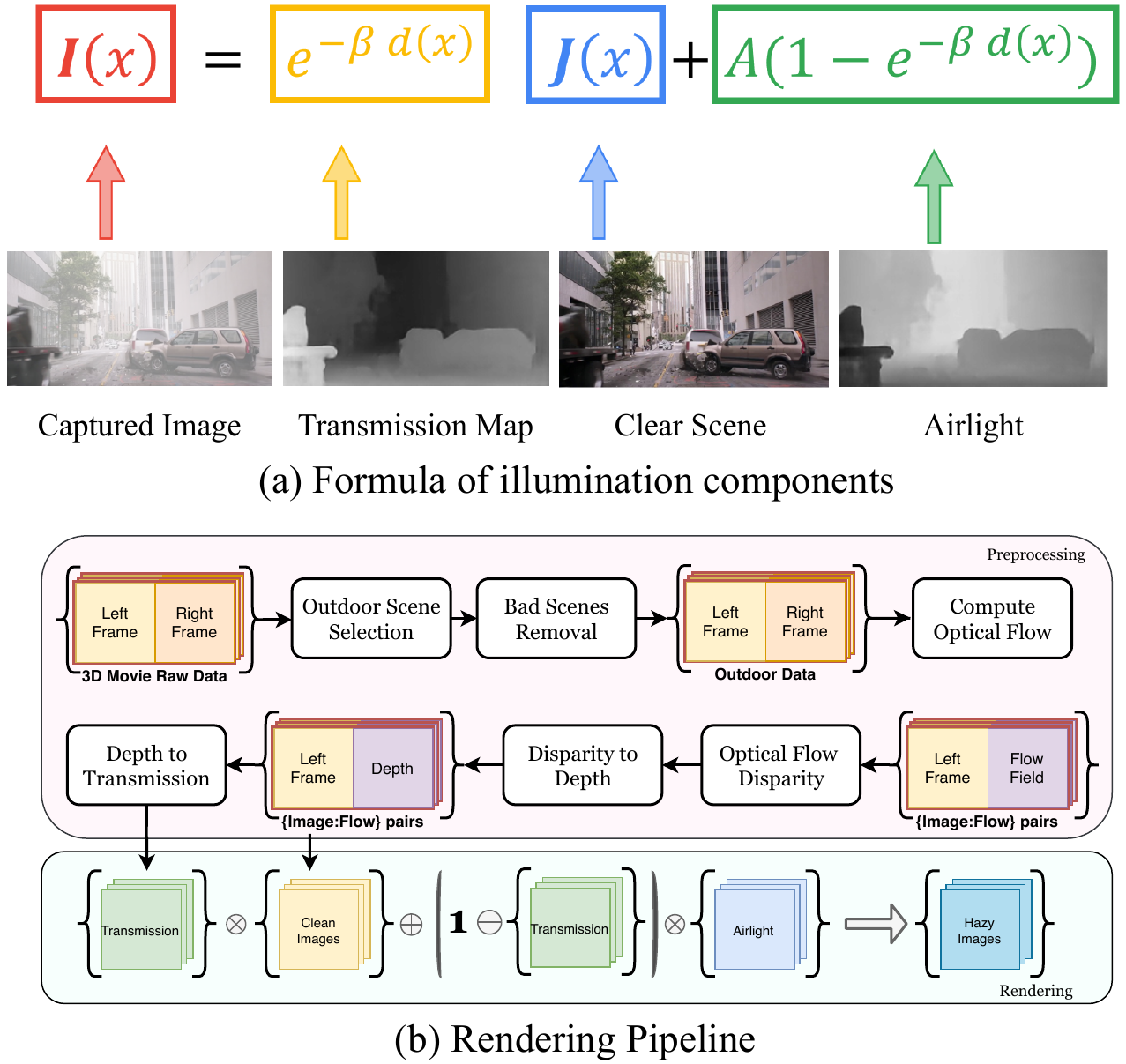}
 \caption{An illustration of the haze image visibility model (a) and the proposed dataset generation pipeline (b). The data rendering pipeline shows all the key stages to transform a raw 3D movie video into a sequence of image-depth pairs. With a series of high-resolution images and their paired depth maps, we create the hazy image data based on the visibility model described in (a). }
 \label{fig:camera}
\vspace{-15pt}

\end{figure}

\section{Large-scale Real Outdoor Dataset}
\label{sec:dataset}
In this section, we introduce the details of our large-scale fine-depth outdoor dataset~\textbf{LSFD}. To elaborate, we first introduce the overall features of the dataset in Sect.~3.1, followed by the motivation to choose high-quality movies as our dataset source. Finally, we describe the details of the 3D movie pre-process in Sect.~3.3 and the haze rendering process in Sect.~3.4. 

\subsection{Dataset Overview}
Table~\ref{tab:datasets} shows the key specifications of our dataset and comparison with existing datasets.  The proposed dehazing dataset contains a total of 10000 images, of which 8000 images are for training set and the rest 2000 images belong to the test set.  The hazy images are rendered on 2000 high-quality clean images, extracted from a series of High-Definition (HD) 3D movies. In order to diversify the background scenes, we extract more than 100 key stereo frame pairs from each of the 22 movies, resulting more than 2400 different background scenes. After our pre-process pipeline, we kept final 2000 high-quality background scenes. For each background scene, we render 5 hazy images on each clean background frame with various haze densities. For the training set, each sample contains four items including the rendered hazy image $\mathbf{I}$, the transmission map $\mathbf{T}$, the atmospheric light value $\mathbf{A}$ and the corresponding clean background image $\mathbf{J}$.

\subsection{Outdoor Image with Fine Depth}
\footnote{We have listed the comparison of all the existing general dehazing datasets in Table~\ref{tab:datasets}.} Sakaridis \cite{sakaridis2018semantic} dataset focuses specifically on semantic scene understanding for the driving scenes, which is not listed.
From Table~\ref{tab:datasets}, we can see that early datasets contain very limited number of samples, which are not suitable for modern data-driven methods.
Existing large-scale outdoor dehazing dataset, \ie~\ the "RESIDE-SOTS" dataset ~\cite{li2019reside}, however is rendered based on the monocular depth estimation method~\cite{liu2015learning}.  Monocular depth estimation is not physically valid in terms of providing ground truth for haze rendering and hence the depth results are not always reliable throughout different outdoor scenes, particularly those depth estimation methods, which they use,  does not perform reliably at many texture-less and ambiguous areas. In contrast we utilizes multi-view stereo \cite{PWC}, which is based on geometric measures and can provide robust and accurate outdoor depth. 

Other than quality of depth and the absolute number of training samples,  a good dataset should have a good coverage of diversified real scenes. So as to enable general applicability and avoid dataset bias. To this end, we find 3D movies provide the largest known source, presenting the possibility of capturing millions of outdoor scenes. Modern high-quality 3D movies are usually captured by top-notch stereo cameras and feature diverse and dynamic environments that range from human-centric imagery (such as Hollywood films) to nature scenes with landscapes and animals in scientific documentaries.  To that end, we select a diverse set of 22 recent 3D movies with stereo image-pairs available.

\begin{figure*}[!t] 
\vspace{-10pt}
 \subfloat{\includegraphics[width=0.33\linewidth]{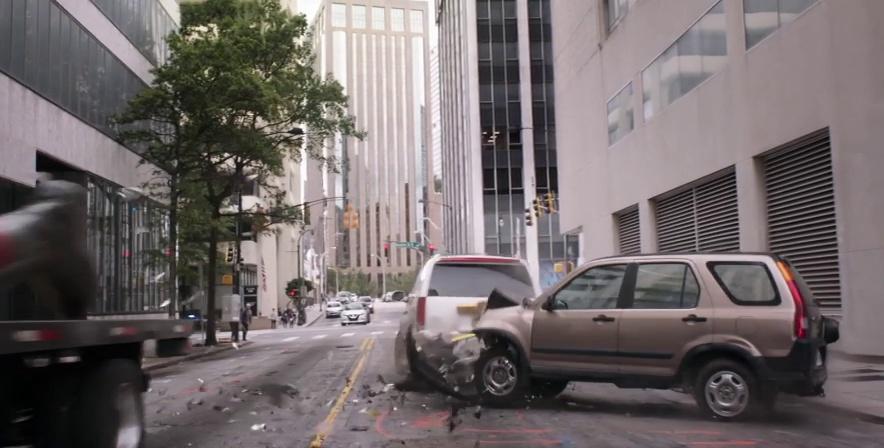}}
 \subfloat{\includegraphics[width=0.33\linewidth]{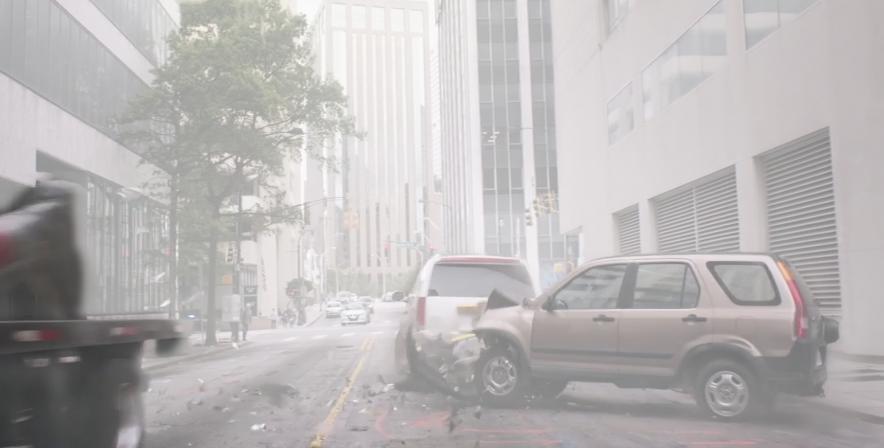}}
 \subfloat{\includegraphics[width=0.33\linewidth]{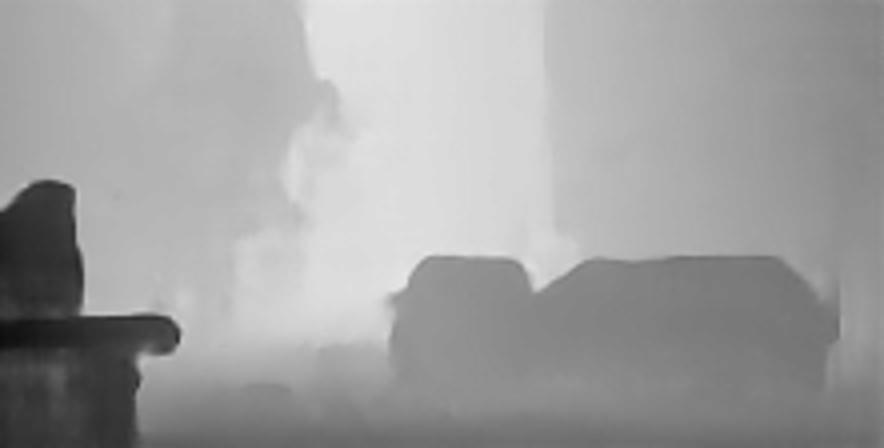}} \vspace{-10pt} \\
 \subfloat{\includegraphics[width=0.33\linewidth]{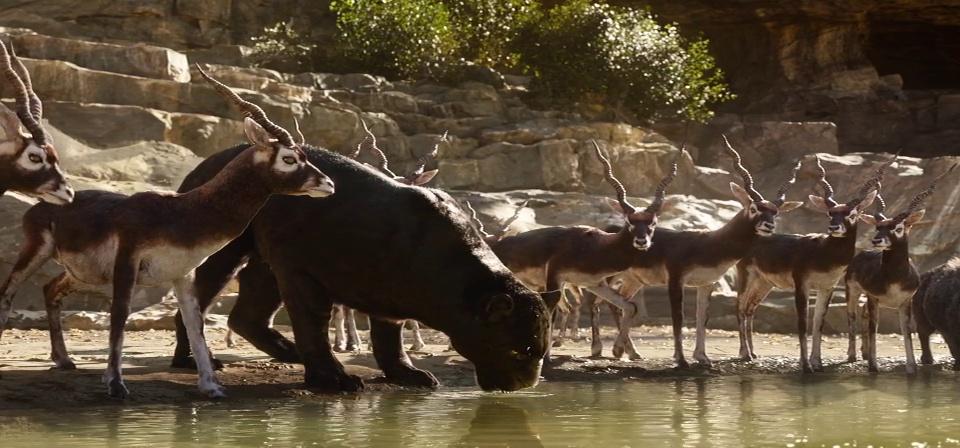}}
 \subfloat{\includegraphics[width=0.33\linewidth]{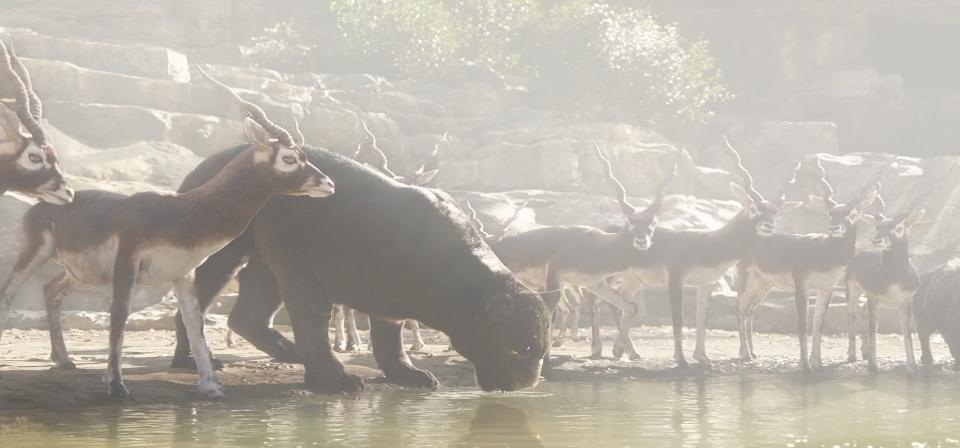}}
 \subfloat{\includegraphics[width=0.33\linewidth]{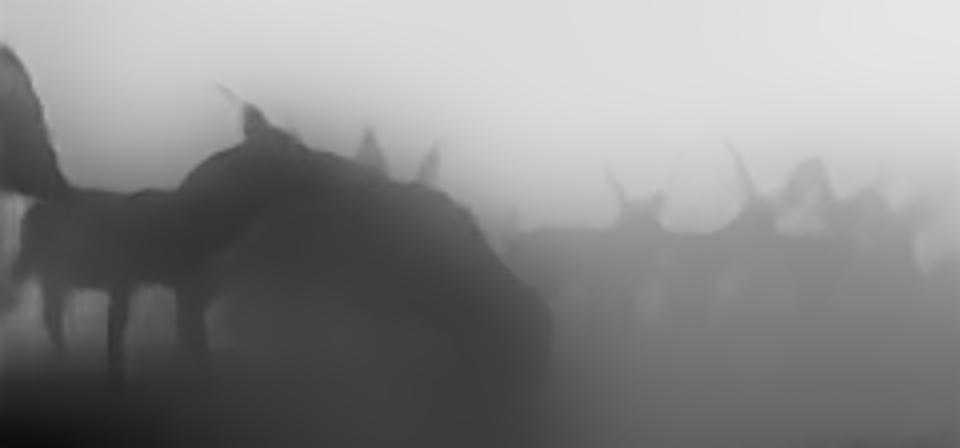}} \vspace{-10pt} \\ 
 \setcounter{subfigure}{0}
 \subfloat[Clean Background (Ground-truth)]{\includegraphics[width=0.33\linewidth]{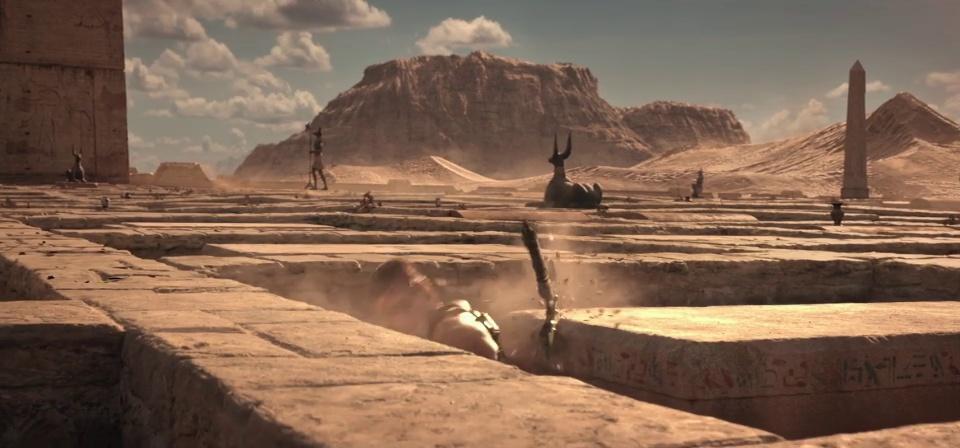}}
 \subfloat[Rendered Hazy Image]{\includegraphics[width=0.33\linewidth]{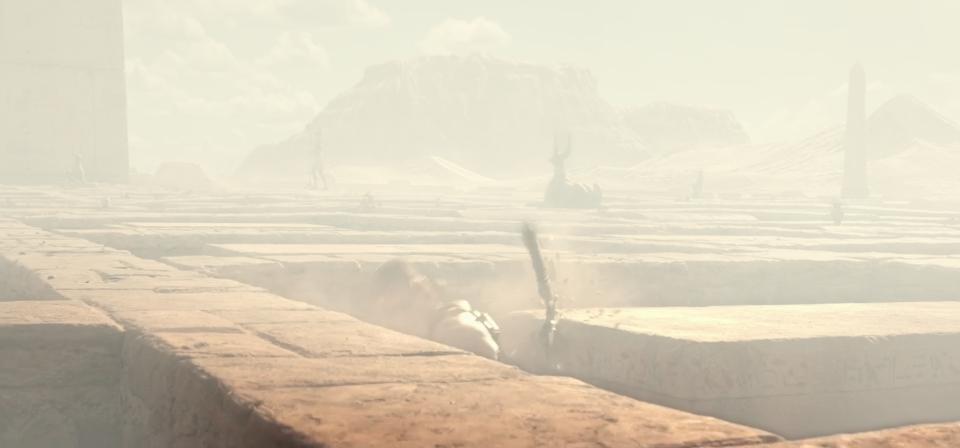}}
 \subfloat[Depth Map]{\includegraphics[width=0.33\linewidth]{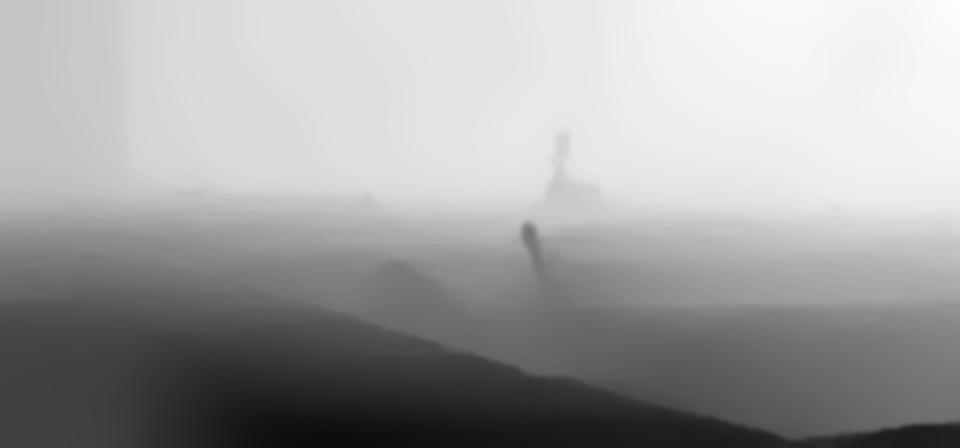}} \\
 \caption{Examples of the proposed dataset, while the clear images are stereo pairs, only one is shown in the figure.}
 \label{fig:existing}
 \vspace{-15pt}
\end{figure*}

\subsection{High Quality Depth using 3D HD Movies}

\noindent \textbf{Outdoor Scene Selection}
In order to obtain rich and diverse outdoor scenes, we carefully select 22 3D-movies. In a typical 2-hour movie, the video may contain up to hundreds of different indoor and outdoor scenes. In order to obtain consecutive frames in the same scene,  we split the long video into individual segments separated by scene transitions. For automatic scene selection, we use scene detection~\cite{pyscenedetect} to split each movie into independent scene. For each segment, we extract up to 2 frames (image pairs) as clean background images. These two frames are selected from the first and the last of the segment respectively to ensure larget displacement. All the extracted frames form the image set $S_0$. Note that $S_0$ contains both indoor and outdoor scenes. We separate the indoor and outdoor scenes using semantic segmentation. We use state-of-the-art method~\cite{zhou2018semantic}. From the result, we only select the images with typical outdoor objects including grassland, road, building, landscape, \etc Lastly, not all the images from $S_1$ are useful for haze rendering because of various reasons. For example, some of the scenes are recorded under foggy, hazy, or smoky conditions. Others may contain strong motion blur or de-focus blur, low-light or dark images, texture-less images, foggy or hazy backgrounds, science-fiction or other style unnatural scenes, \etc We therefore do a manual check and remove the flaw samples. The final set is denoted as $S_2$. 

\begin{figure*}[!t]
    \centering
    \subfloat[Hazy Image \cite{li2019reside}]{\includegraphics[width=0.24\linewidth]{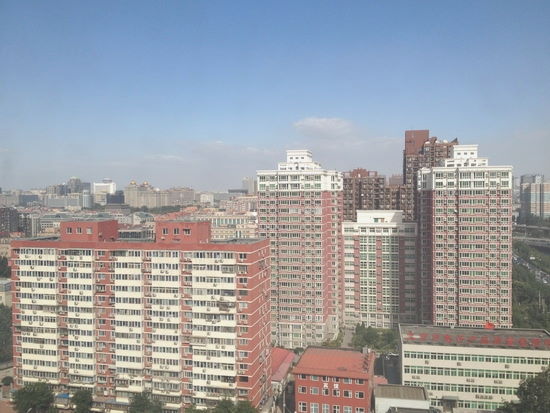}}
    \subfloat[Depth \cite{li2019reside}]{\includegraphics[width=0.24\linewidth]{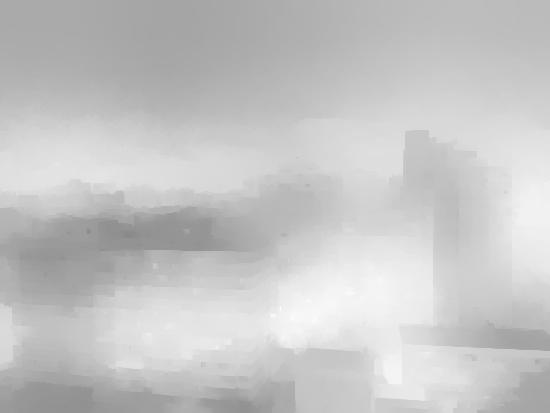}} 
    \subfloat[Hazy Image \cite{li2019reside}]{\includegraphics[width=0.24\linewidth]{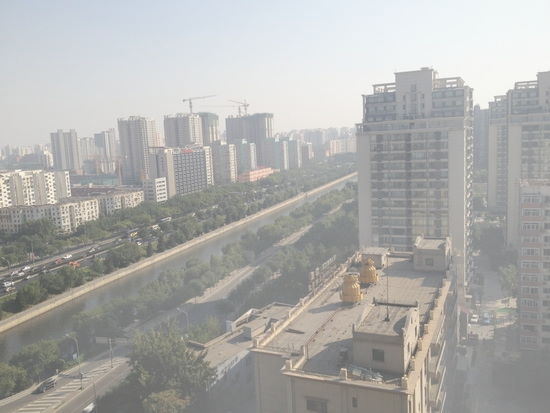}}
    \subfloat[Depth \cite{li2019reside}]{\includegraphics[width=0.24\linewidth]{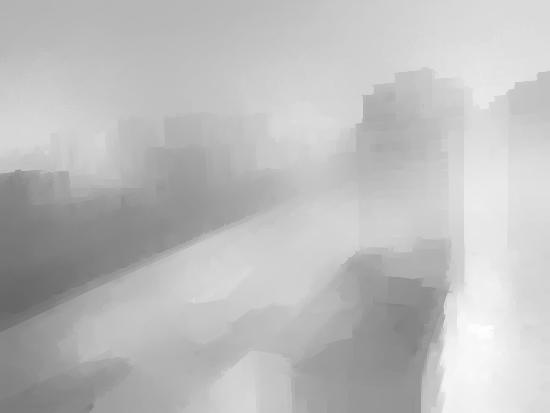}} \\ \vspace{-5pt}
    \subfloat[Our  Hazy Image]{\includegraphics[width=0.24\linewidth]{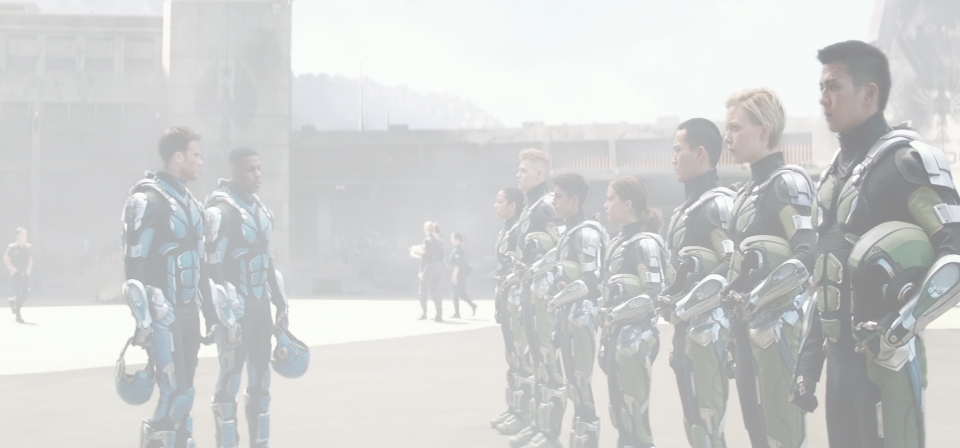}}
    \subfloat[Our Depth]{\includegraphics[width=0.24\linewidth]{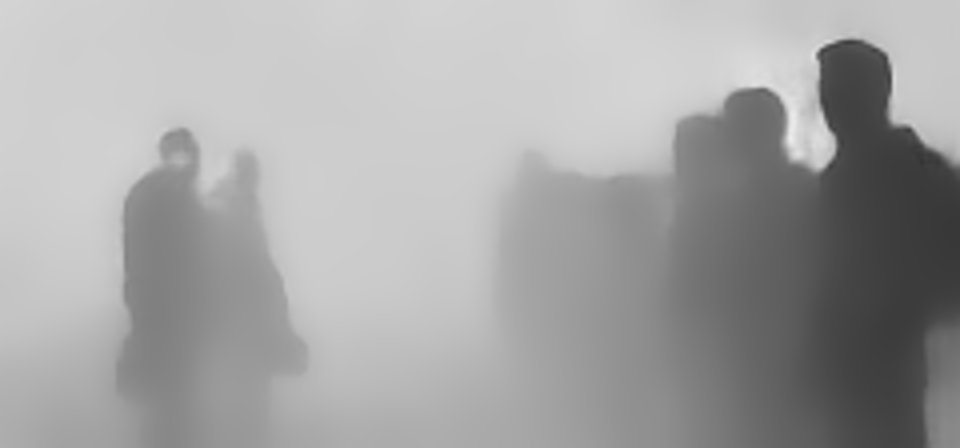}}
    \subfloat[Our  Hazy Image]{\includegraphics[width=0.24\linewidth]{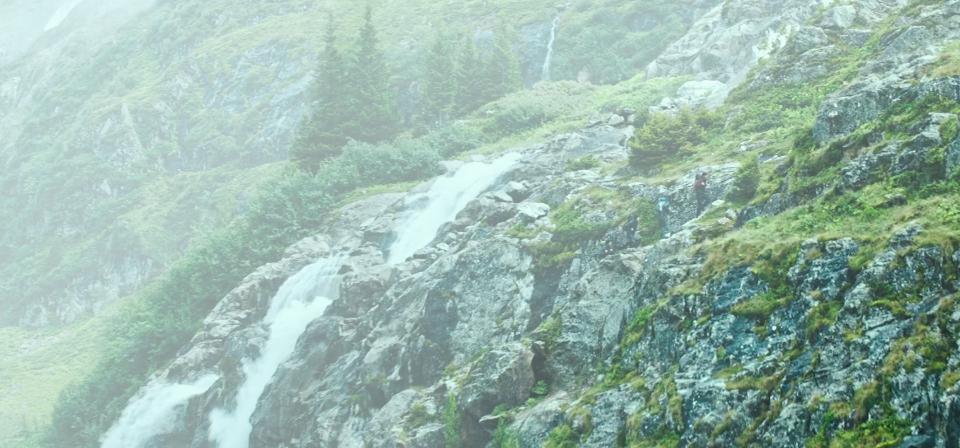}}
    \subfloat[Our Depth]{\includegraphics[width=0.24\linewidth]{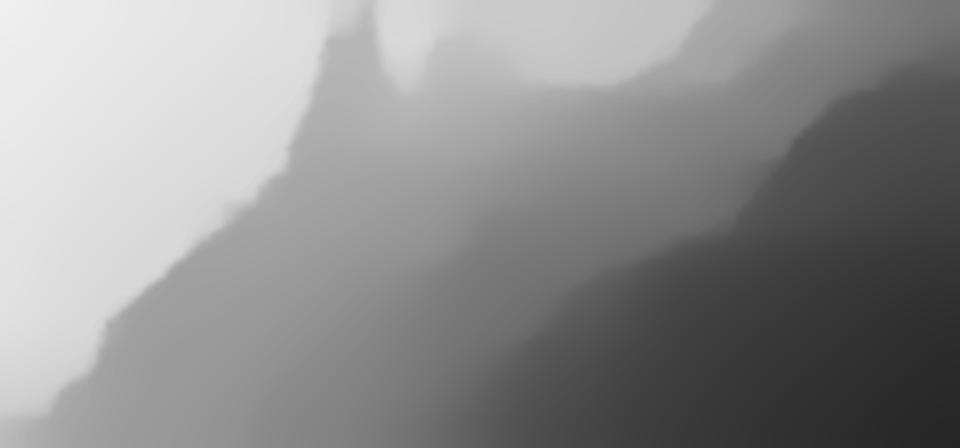}}
    \\ 
    \caption{Hazy images and corresponding depth maps of 
    Comparison on rendering and depth map with RESIDE-OTS \cite{li2019reside}. RESIDE-OTS uses empirical monocular depth estimation ~\cite{liu2015learning}. The unreliable depth also results in unrealistic haze rendering. 
    In contrast, our depth is based on physical measurement and thus reliable; and leads to realistic haze rendering.
    }
    \label{fig:depth_compare_reside}
    \vspace{-20pt}
\end{figure*}

\noindent \textbf{High-Quality Depth Map}
As argued, a major advantage of our proposed dataset is the high-quality depth.
In detail, we ensure that our depth maps are strictly from real measurements. Notice that some of the 3D movies are actually created by synthetic depth, those movie cannot provide reliable depth following the physics model. Therefore, we only use the images which are captured by real stereo cameras. 

In practice, we solve some technical challenges. First, we notice that there are information loss such as camera focal lengths, camera baselines, and intrinsic parameters. Second, while most stereo algorithms assume transition between cameras, which generate only positive disparities, 3D movie data may have negative disparities depending on the camera positioning. 
To solve it,  we use optical flow estimation PWCNet \cite{PWC} instead of applying stereo algorithms. 
We use only the horizontal part of the estimate flow and regard it as disparity as shown in Fig.~\ref{fig:camera}(b) (Middle Row).
Lastly, PWCNet \cite{PWC} produces flow maps at a quarter of the natural resolution. We apply FGI \cite{Li2016fgi} algorithm to up-sample the flow to its original resolution. The benefit of FGI it that it is content aware and can provide sharp and precise boundaries. After obtaining the disparities, the depth is inversely proportional to it.  Fig.~\ref{fig:depth_compare_reside} is a comparison of the depth with RESIDE dataset. As can-be seen, while RESIDE depth can be incorrect and blurry, our depth is more reliable and sharp.

Although 3D movies provide stereo image pairs for every frame, the movie data comes with its own issues as well. First, some 3D films are shot with monocular cameras and the stereo effects are manually added by post-processing. In this case, we do not include those movies in our dataset as we only select movies that were shot using physical stereo cameras. Second, we only select high-definition movies in \textit{Blu-ray} storage format, which allow us to extract high-resolution image pairs.  In addition, camera focal lengths, stereo camera baselines, and camera intrinsic parameters are usually unknown and these parameters vary from one movie to another. It is difficult to directly compute the stereo disparity solely from the image pairs using the state-of-the-art stereo estimation algorithms. In addition, most of the stereo algorithms are designed and trained to estimate disparity in positive ranges only. However, the 3D movie data may contain negative disparity values because of the camera position settings in the stereo rig configuration.  Instead of applying stereo algorithms, we apply the state-of-the-art optical flow estimation method PWCNet \cite{PWC} to handle the positive and negative disparity values. We only retain the horizontal component of the flow field as the disparity as shown in Fig.~\ref{fig:camera}(b) (Middle Row). As the depth is inversely proportional to the horizontal disparity computed in last step, we convert the disparity map to depth map for each image as shown in Fig.~\ref{fig:existing} (a) and (c).

\noindent \textbf{Depth Map Refinement}
The PWCNet \cite{PWC} produces flow fields in a quarter of the original image size.  A bi-linear up-sampling post-processing is usually used to up-sample the flow fields to the full size. In our case, the bi-linear up-sampling usually makes the object's boundary blurry,  which also leads to halo effect on the rendered object's boundary. To that end, we apply FGI \cite{Li2016fgi} algorithm to up-sample the flow fields based on the boundary, contours, and edges of the input image shown in Fig.~\ref{fig:depth_compare_reside}. From the figure, We can observe that the depth discontinuities are more aligned with the actual objects' boundaries in the input images. Hence, the halo effects of the FGI up-sampled results are significantly reduced compared with  depth maps from RESIDE-$\beta$ dataset. 

\subsection{Hazy Image Rendering}
\paragraph{Attenuation}
Our rendering follows the haze model 
We follow Eq.~\ref{eq:scattering_model} to render haze using depth.
Algorithm~\ref{alg:pipeline} describes the details.
We also apply refinement to improve from Eq.~\ref{eq:scattering_model}. 
Background blurriness is a function related to the depth due to Rayleigh scattering,
following McCartney\cite{mccartney1976optics}, we apply Gaussian smooth with depth-aware kernel to the haze-free images. 
To diversify haze density ($\beta$ in Eq.~\ref{eq:scattering_model}), we synthesize 5 hazy images using different $\beta$ which is randomly sampled from a uniform distribution in the interval of $[1.0, 2.5]$.

\paragraph{Atmospheric Light}
The atmospheric light needs to be consistent with the environment lighting and tone of the haze-free image.
We use classic MaxRGB illumination estimation\cite{Land1977TheRT} to determine $\mathbf{A}$:
\begin{equation}
    \centering
    \mathbf{A}_M = \max_{\mathbf{x}\in{\mathbf{\Omega}}}{M(\mathbf{x})}, M \in \{ R, G, B \},
    \label{eq:airlight_rendering}
\end{equation}
where $R, G, B$ indicate red, green, blue channel of a haze-free image.
$\mathbf{\Omega}$ means all pixels of $\mathbf{I}$.  

\paragraph{Splitting}
In total, we have 2000 haze-free images which now generates in total 10,000 hazy images.
To training and testing split, the training set contains 8000 hazy images (1600 haze-free) and the test set contains 2000 hazy images (400 haze-free).
Examples are shown in Fig.~\ref{fig:existing}.

\vspace{-10pt}
\begin{algorithm}[t]
\caption{Proposed Haze Rendering Algorithm}
\label{alg:pipeline}
\begin{algorithmic}[1]
\State \textbf{Input: }Clean Image $\mathbf{J}$ and its depth map $\mathbf{d}$
\State $\mathbf{J}_{blur}(\mathbf{x}) = \mathsf{imgaussfilt}(\mathbf{J}(\mathbf{x}), \mathbf{\sigma}(\mathbf{x}))$. The kernel varies according to depth: $\mathbf{\sigma}(\mathbf{x}) = 1.5  \mathbf{d}(\mathbf{x})$.
\State Obtain Transmission $\mathbf{T} = \exp^{-\beta  \mathbf{d}}, \beta \sim \textit{U}(1.0, 2.5)$.
\State Obtain global atmospheric light  $\mathbf{A}$ based on Eq.~\ref{eq:airlight_rendering}.
\State \textbf{Output: } Hazy Image $\mathbf{I} = \mathbf{T}_{blur} \mathbf{J}  + (1 - \mathbf{T}_{blur}) \mathbf{A}$. 
\end{algorithmic}
\end{algorithm}
\vspace{-5pt}

\begin{table}[!tb]
    \centering
	\vspace{-5pt}
    \caption{The quantitative results of the state of the art methods tested on the proposed dataset \textbf{LSFD} \textit{test} set as well as on RESIDE-SOTS dataset. For CNN-based methods, we use the pre-trained weights provided by the authors, indicated as 'official'. For those methods without official weights, we train them with RESIDE-OTS and LSFD and mark the training sets after each method respectively.  }
    \setlength{\tabcolsep}{2pt}
    \begin{tabular}{l|cccc|cc|ccc}
	  {Methods} &\multicolumn{4}{|c|}{LSFD}  &\multicolumn{2}{c|}{RESIDE-SOTS} & Time (s) & Platform \\
	  \hline
         { } &  PSNR & PSNR-STD & SSIM & SSIM-STD & PSNR & SSIM &  &  \\
        \toprule
        BCCR \cite{meng2013bccr}    & 17.91 & 2.27 &  0.729 & 0.083 & 16.88 & 0.7913 & 3.85 & CPU\\
        DCP \cite{he2011dcp}        & 19.56 & 2.38  & 0.761 & 0.069  & 16.62 & 0.8179 & 1.62 & CPU\\
        CAP \cite{zhu2015cap}       & 14.64 & 3.79 & 0.715  & 0.111 & 19.05 & 0.8364 & 0.95 & CPU\\
        Non-Local \cite{berman2016nonlocal} & 18.06 & 2.78 & 0.727 & 0.079 & 17.29 & 0.7489 & 9.89 & CPU \\
        GRM \cite{chen2016grm}      & 17.48 & 2.27& 0.682  & 0.083 & 18.86 & 0.8553 & 83.96 & CPU\\
        \midrule
        MS-CNN(official) \cite{ren2016mscnn}  & 10.85 & 3.17 & 0.687 & 0.110 & 17.57 & 0.8102 & 2.60 & GPU\\
        AODnet(official) \cite{li2017aod}     & 12.42 & 2.99 & 0.639 & 0.099 & 19.06 & 0.8504 & 0.65 & GPU\\
        DehazeNet(official) \cite{cai2016dehazenet} &  19.76 & - & 0.788 & - &  21.14 & 0.8472 & 2.51 & GPU\\
        DCPDN(official) \cite{zhang2018dcpdn}  & 15.38 & - & 0.660  & -  & - & - & - & - \\
        GridDehazeNet(official) \cite{liu2019griddehazenet} & 12.37 & 3.38 & 0.659 & 0.120  & 30.86 & 0.9819 & 0.26 & GPU \\
        EPDN(official) \cite{Qu_2019_CVPR} & 14.44 & 3.26 & 0.797 & 0.081 & 25.06 & 0.9232 & 0.80 & GPU\\
        4K Dehazing(official) \cite{Zheng_2021_CVPR} & 17.31 & 3.01 &  0.727 & 0.090 & 18.39 & 0.882 & \textbf{0.03} & GPU \\
        PSD(official) \cite{Chen_2021_CVPR} & 10.79 & 2.53 & 0.619 & 0.113 & 15.15 & 0.737 & 0.93 & GPU \\
        MSBDN(official) \cite{MSBDN-DFF} & 25.99 & \textbf{2.02} & 0.870 & \textbf{0.044} & 33.79 & 0.9835  & 0.11 & GPU\\
        FFA-Net(official) \cite{FFANet} & 25.37 & 3.25 & 0.941 & 0.074 & 33.57  & 0.9840 & 0.23 & GPU\\
        DF2M-Net \cite{deng2019deep} (RESIDE) & 24.66 & 4.23 & 0.918 & 0.097 & \textbf{34.29} & \textbf{0.9844} & 0.07 & GPU \\
        DF2M-Net \cite{deng2019deep} (LSFD) & \textbf{28.49} & 3.75 & \textbf{0.943} & 0.083 & 29.88 & 0.9376 & 0.07 & GPU \\
        \bottomrule
    \end{tabular}
    \label{tab:ourdata_quant}
    \vspace{-10pt}
\end{table}

\section{Benchmark}
\label{sec:exp}
\paragraph{Experiment Setup}
We provide systematical benchmark on 16 representative conventional methods and data-driven methods.
The traditional methods include Boundary Constrained Context Regularization (BCCR) \cite{meng2013bccr}, Color Attenuation Prior (CAP) \cite{zhu2015cap}, Dark-Channel Prior (DCP) \cite{he2011dcp}, Artifact Suppression via Gradient Residual Minimization (GRM) \cite{chen2016grm}, and Non-local Image Dehazing (non-local) \cite{berman2016nonlocal}. These methods do not require training process and therefore we directly run the methods on the dedicated test sets respectively.
Data-driven methods include DehazeNet \cite{cai2016dehazenet}, Multi-scale CNN (MSCNN) \cite{ren2016mscnn}, All-in-One Dehazing (AODNet) \cite{li2017aod}, Densely Connected Pyramid Dehazing Network (DCPDN) \cite{zhang2018dcpdn}, grid  \cite{liu2019griddehazenet}, Enhanced Pix2pix Dehazing Network (EPDN) \cite{Qu_2019_CVPR}, Multi-Scale Boosted Dehazing Network with Dense Feature Fusion (MSBDN) \cite{MSBDN-DFF}, DF2M-Net \cite{deng2019deep}, 4K Dehazing \cite{Zheng_2021_CVPR}, FFA-Net \cite{FFANet} and  PSD \cite{Chen_2021_CVPR}.

We use PSNR \cite{PSNR} and SSIM \cite{SSIM} metrics to evaluate all the methods. The conventional methods are directly applied on the proposed test set. For CNN-based methods, we directly use the weights provided by the author to test on the proposed datasets if available. For those methods which do not lease any pre-trained weights, we particularly train the methods on the training sets and evaluate them on the test set.  

\begin{figure*}[!ht]
\centering
 \subfloat[Input]{\includegraphics[width=0.24\linewidth]{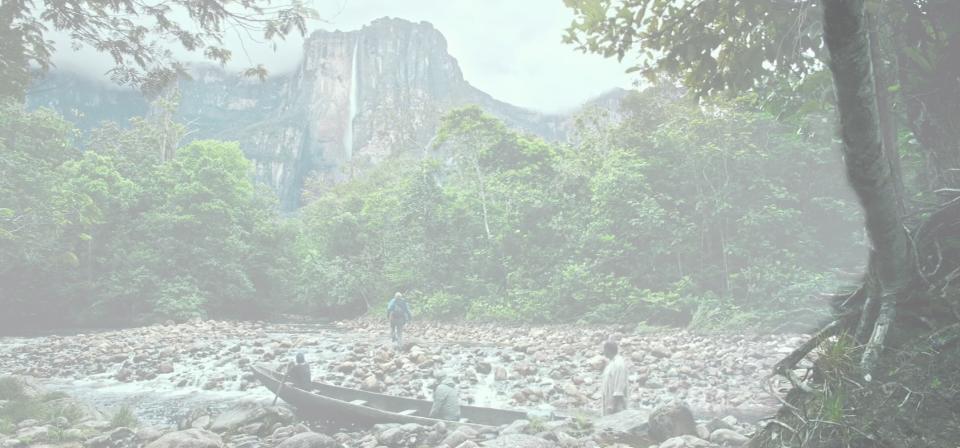}}
 \subfloat[CAP \cite{zhu2015cap}]{\includegraphics[width=0.24\linewidth]{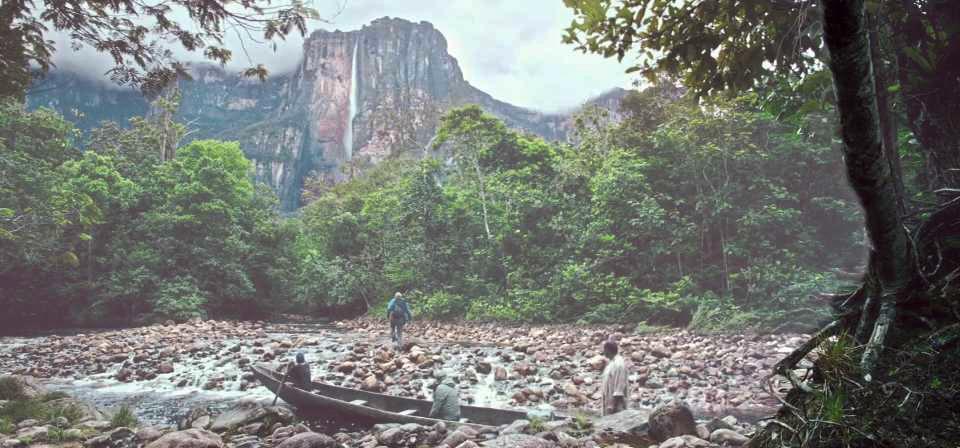}}
 \subfloat[DCP \cite{he2011dcp}]{\includegraphics[width=0.24\linewidth]{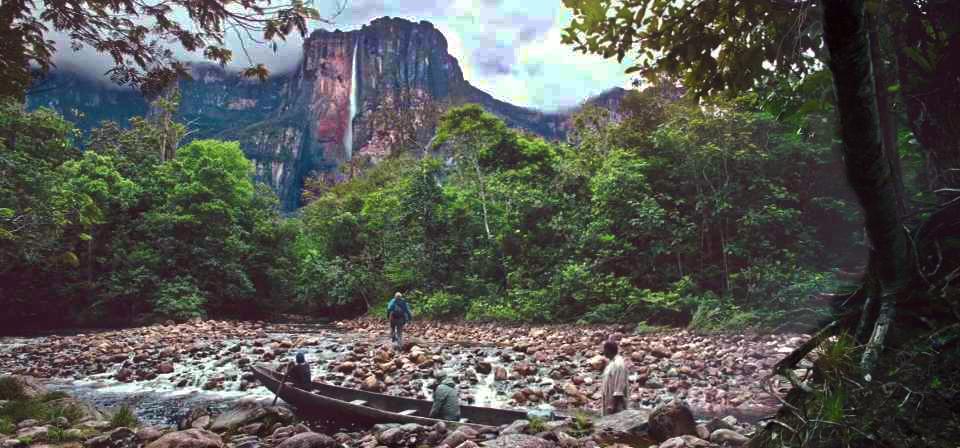}}
 \subfloat[Non-Local \cite{berman2016nonlocal}]{\includegraphics[width=0.24\linewidth]{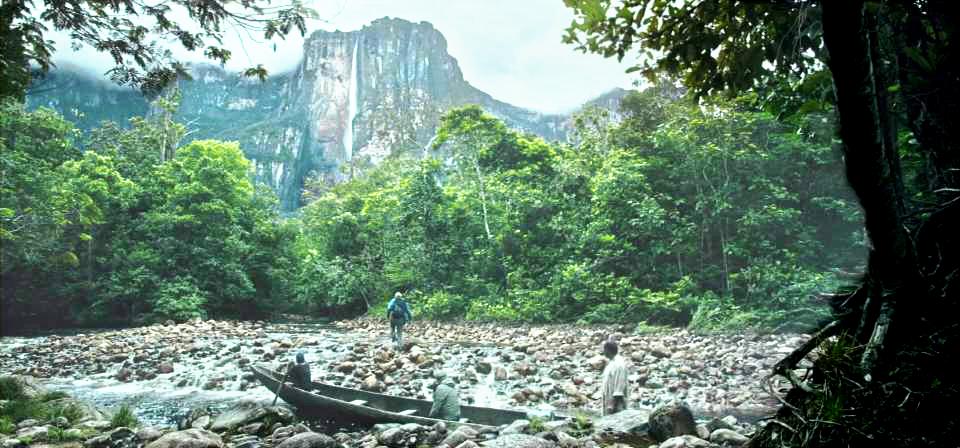}}  \\ \vspace{-10pt}
 \subfloat[MS-CNN \cite{ren2016mscnn}]{\includegraphics[width=0.24\linewidth]{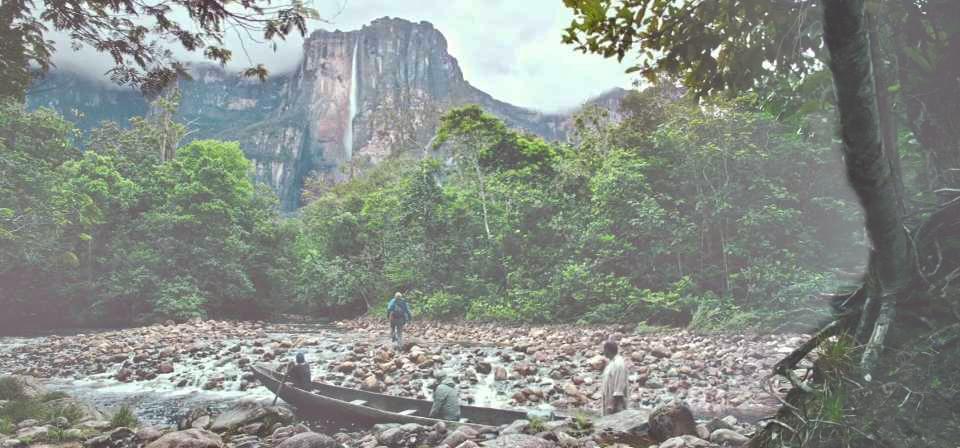}} 
 \subfloat[AODNet \cite{li2017aod}]{\includegraphics[width=0.24\linewidth]{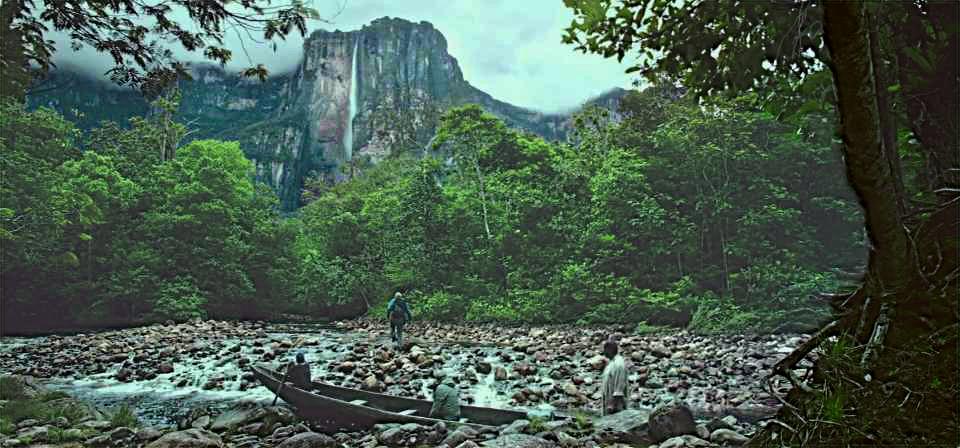}}
 \subfloat[{\scriptsize GridDehazeNet}~\cite{liu2019griddehazenet}]{\includegraphics[width=0.24\linewidth]{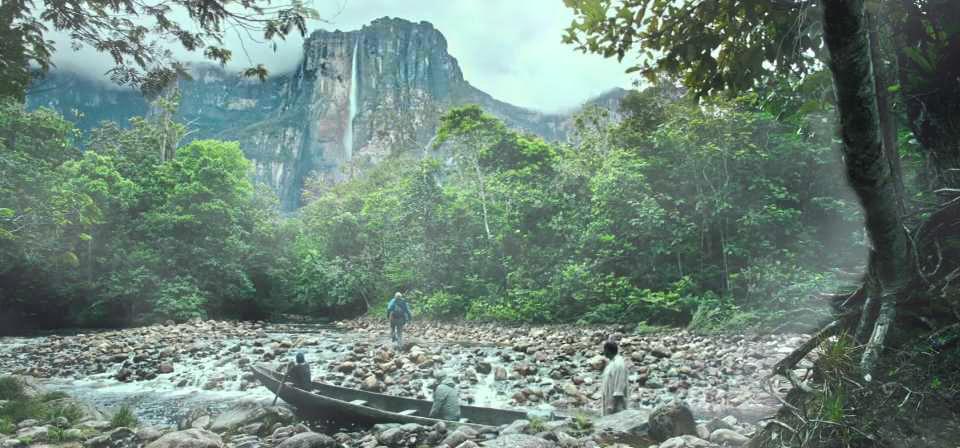}}  
 \subfloat[MSDBN\cite{zhu2015cap}]{\includegraphics[width=0.24\linewidth]{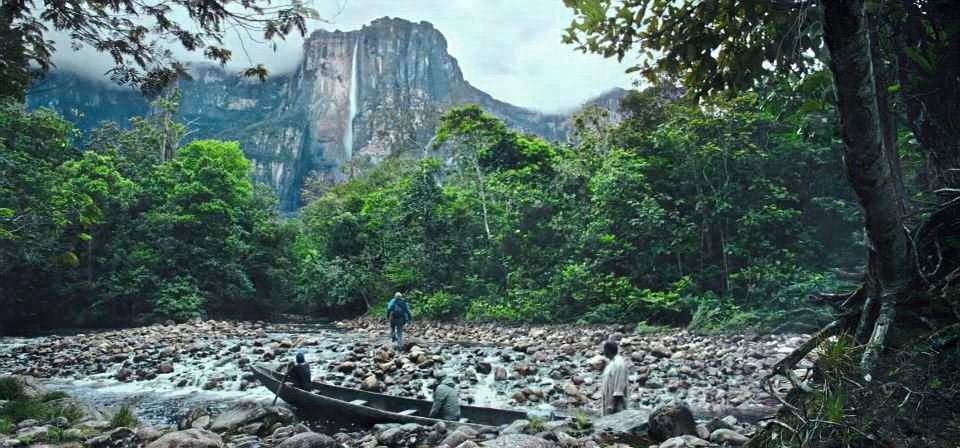}}   \\  \vspace{-10pt}
 \subfloat[FFA-Net \cite{zhu2015cap}]{\includegraphics[width=0.24\linewidth]{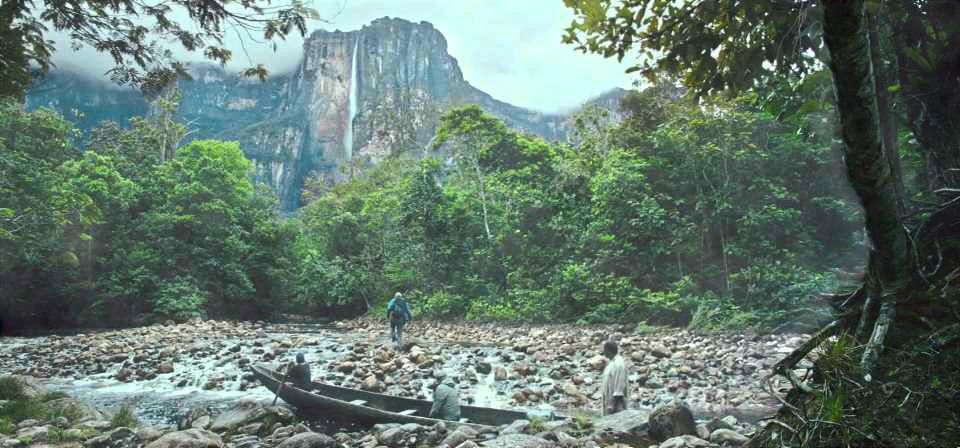}}
 \subfloat[DehazeNet \cite{cai2016dehazenet}]{\includegraphics[width=0.24\linewidth]{fig/qualitative/set5/dcpdn-M13-PointBreak_seg_0885_123549_l.jpg}}
 \subfloat[DCPDN \cite{zhang2018dcpdn}]{\includegraphics[width=0.24\linewidth]{fig/qualitative/set5/dcpdn-M13-PointBreak_seg_0885_123549_l.jpg}}
 \subfloat[GT]{\includegraphics[width=0.24\linewidth]{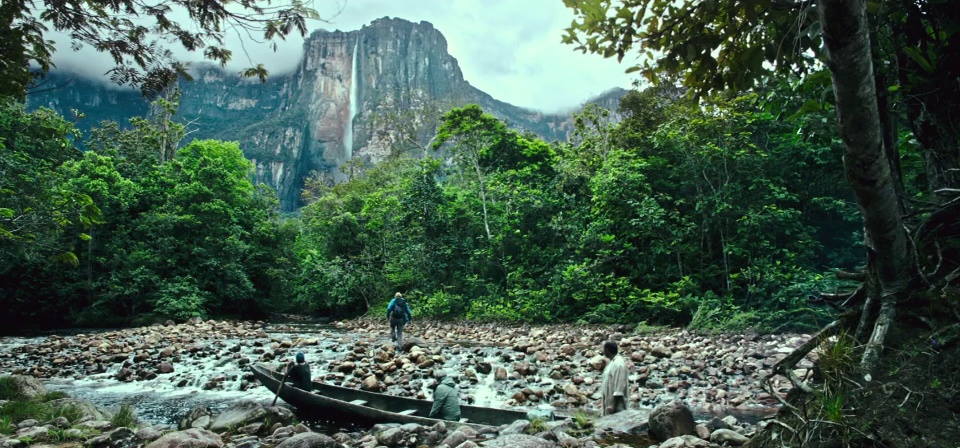}} 
 \caption{Qualitative results on the proposed \textbf{LSFD} dataset. }
 \label{fig:proposed1}
\vspace{-15pt}
\end{figure*}

\begin{figure*}[!ht]
\centering
 \subfloat[Input]{\includegraphics[width=0.24\linewidth]{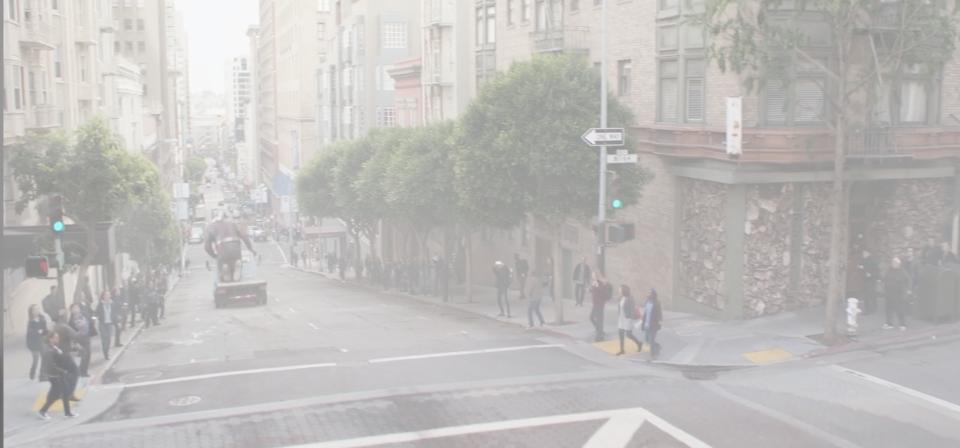}}
 \subfloat[CAP \cite{zhu2015cap}]{\includegraphics[width=0.24\linewidth]{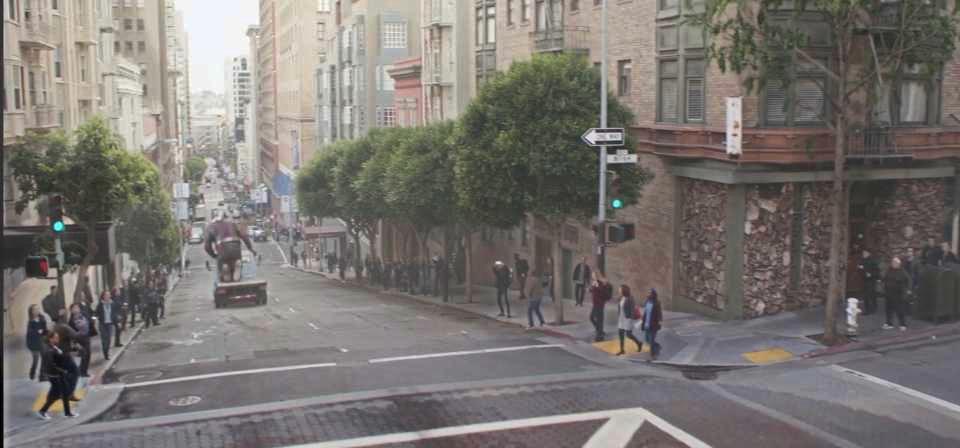}}
 \subfloat[DCP \cite{he2011dcp}]{\includegraphics[width=0.24\linewidth]{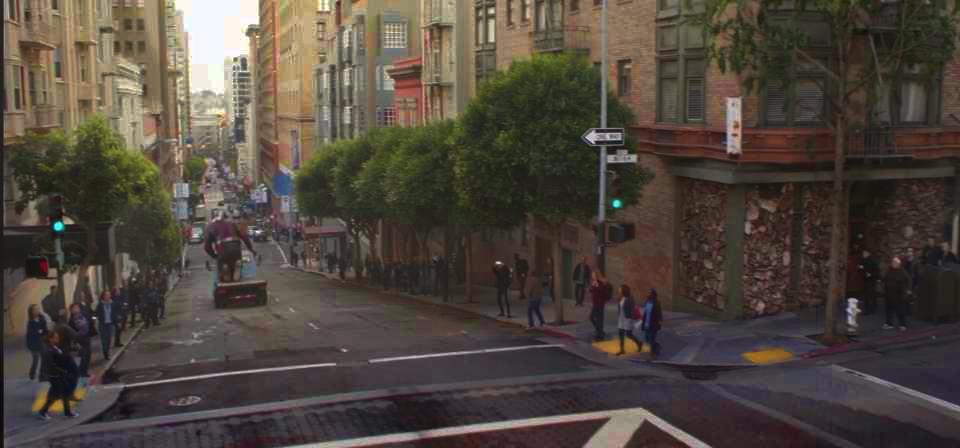}} 
 \subfloat[Non-Local \cite{berman2016nonlocal}]{\includegraphics[width=0.24\linewidth]{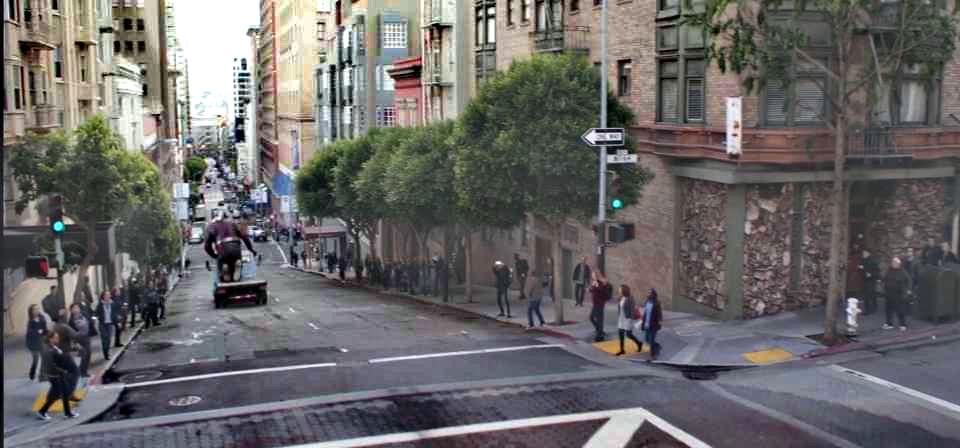}}  \\    \vspace{-10pt}
 \subfloat[MS-CNN \cite{ren2016mscnn}]{\includegraphics[width=0.24\linewidth]{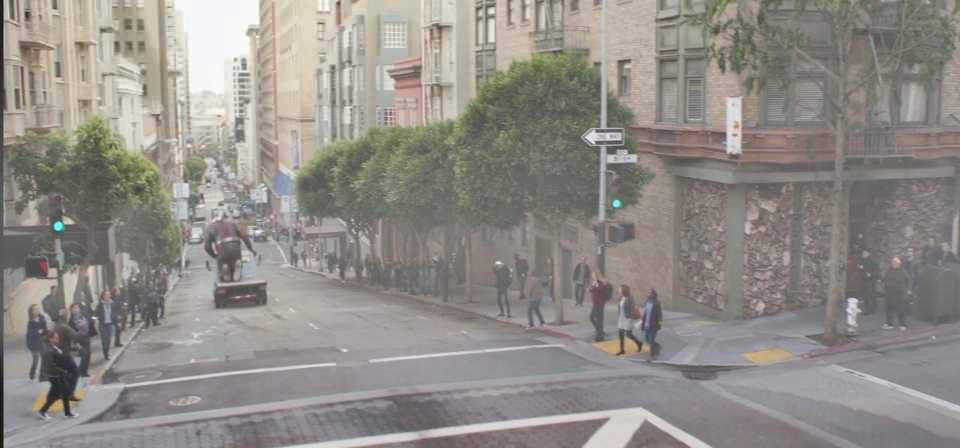}} 
 \subfloat[AODNet \cite{li2017aod}]{\includegraphics[width=0.24\linewidth]{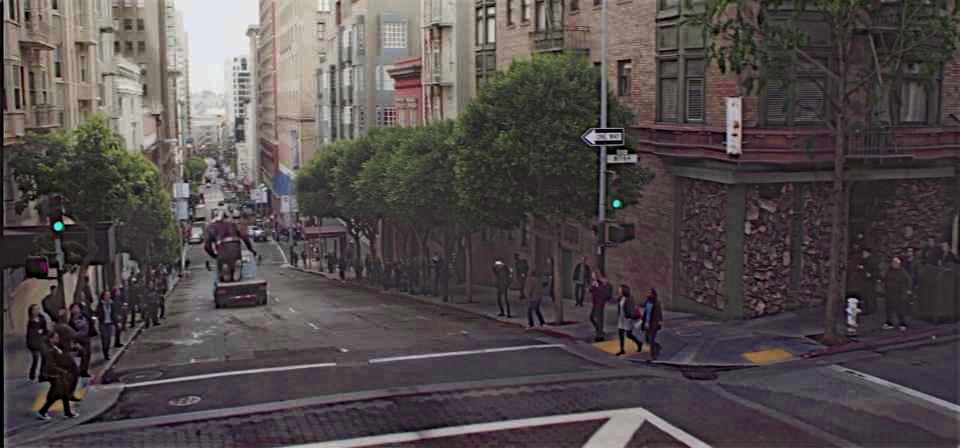}}
 \subfloat[\scriptsize GridDehazeNet \cite{liu2019griddehazenet}]{\includegraphics[width=0.24\linewidth]{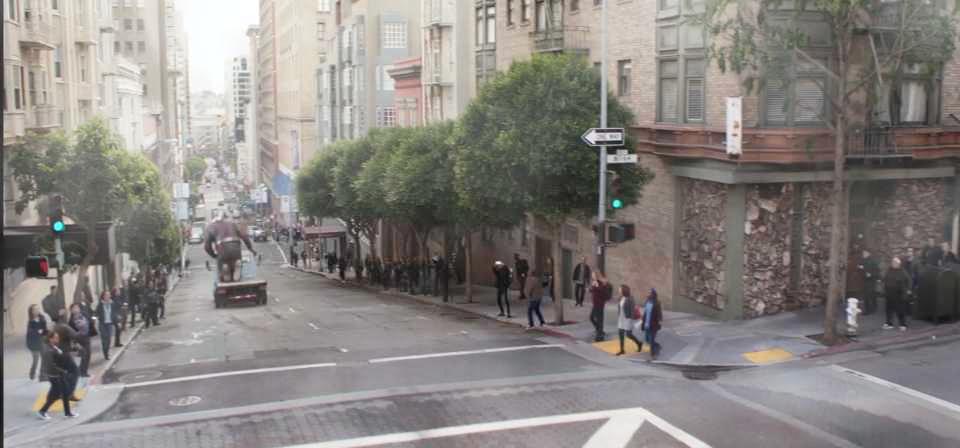}}  
 \subfloat[DehazeNet \cite{cai2016dehazenet}]{\includegraphics[width=0.24\linewidth]{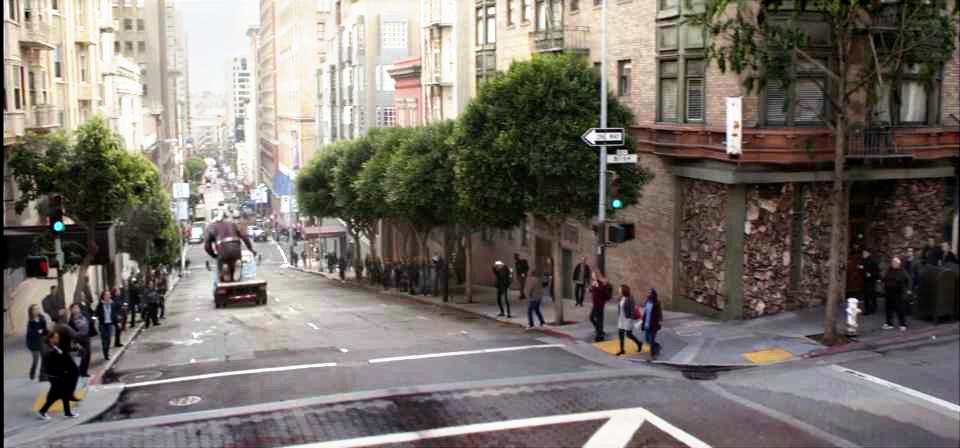}} \\    \vspace{-10pt}
 \subfloat[FFA-Net \cite{ren2016mscnn}]{\includegraphics[width=0.24\linewidth]{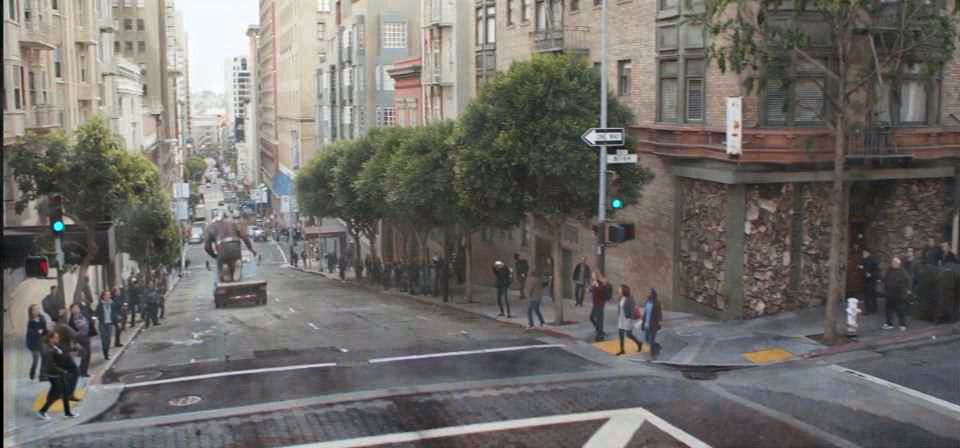}}  
 \subfloat[MSBDN \cite{ren2016mscnn}]{\includegraphics[width=0.24\linewidth]{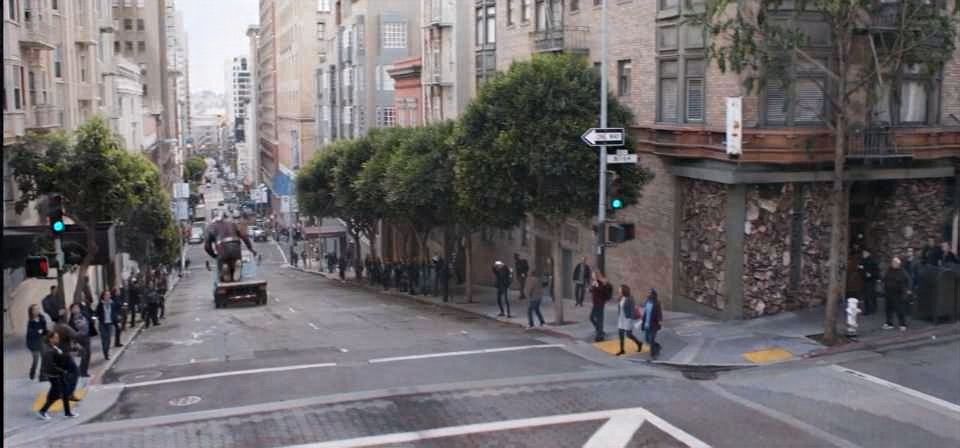}}
 \subfloat[DCPDN \cite{zhang2018dcpdn}]{\includegraphics[width=0.24\linewidth]{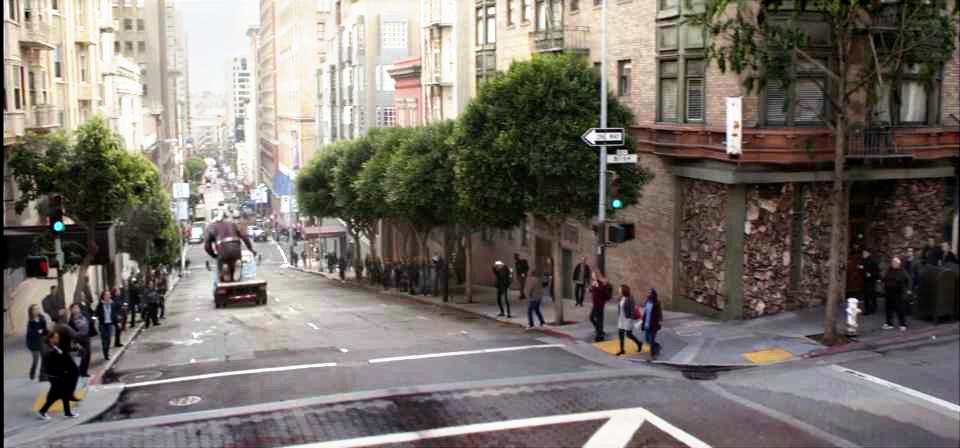}}
 \subfloat[GT]{\includegraphics[width=0.24\linewidth]{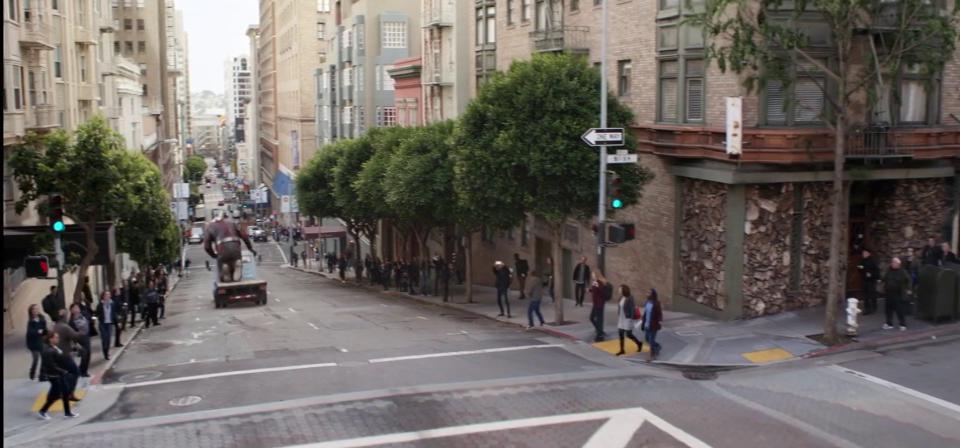}} 
 \caption{Results comparison on the proposed \textbf{LSFD} dataset. }
 \label{fig:proposed2}
\vspace{-15pt}
\end{figure*}

\subsection{Results and Analysis}
\paragraph{Overall performance}
First of all, we provide an overview of the performance of all the selected dehazing algorithms in Table~\ref{tab:ourdata_quant}.
For reference, we also benchmark the methods on RESIDE-$\beta$ dataset. The numbers are quoted from their benchmark if  available, otherwise we benchmark them if they release training and testing code. Time and platform is also provided, as extra information.
From the results, we can see that the PSNR and SSIM is not as saturated as RESIDE-$\beta$, which shows that the proposed dataset it more diversified and less likely to be over-fitted. In addition, we see performance ranking on two different datasets are clearly different, which shows the significant different of two datasets, especially the quality of depth.

\begin{figure*}[!ht]
\centering
\vspace{-15pt}
 \subfloat{\includegraphics[width=0.24\linewidth]{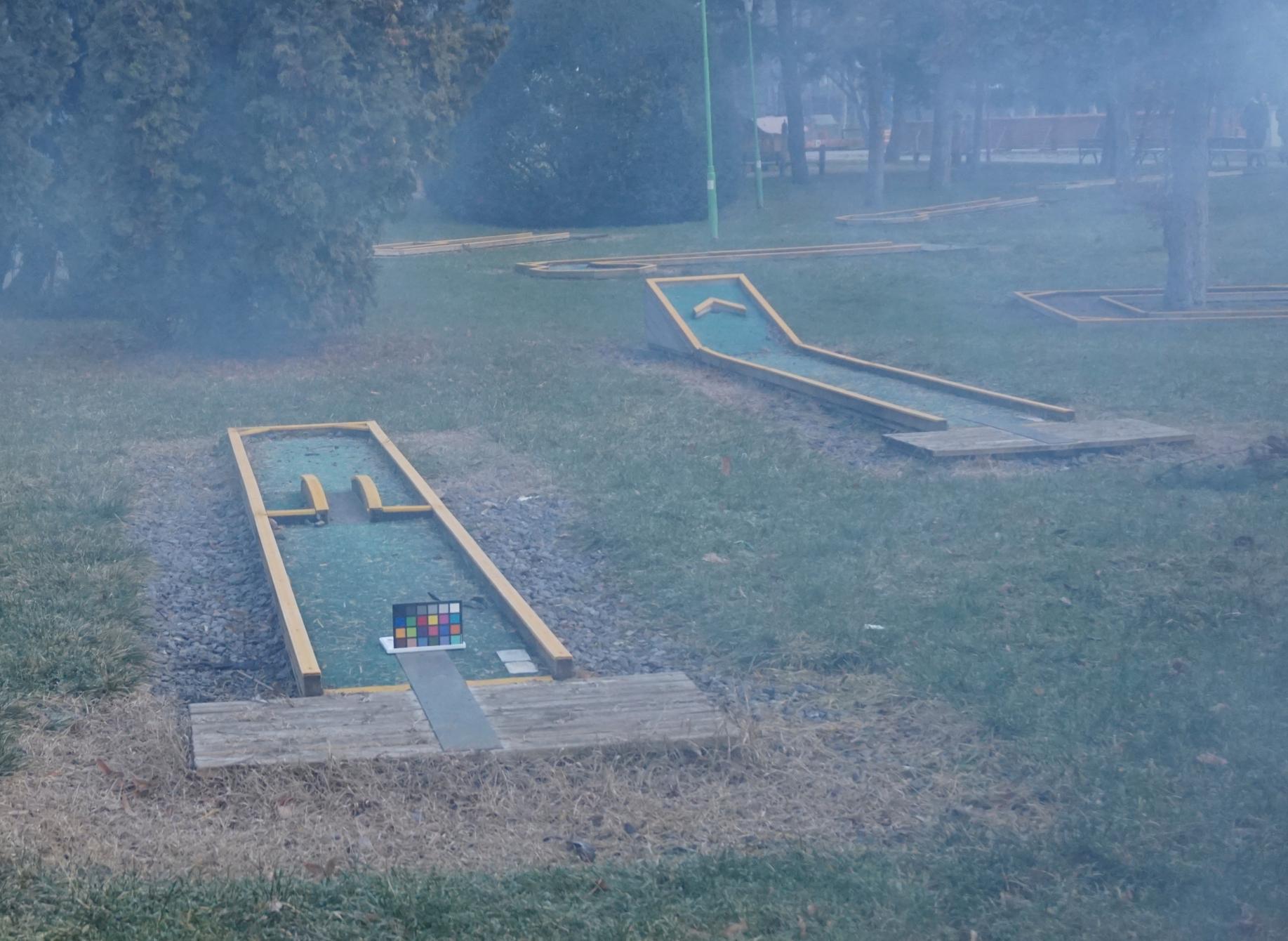}}
 \subfloat{\includegraphics[width=0.24\linewidth]{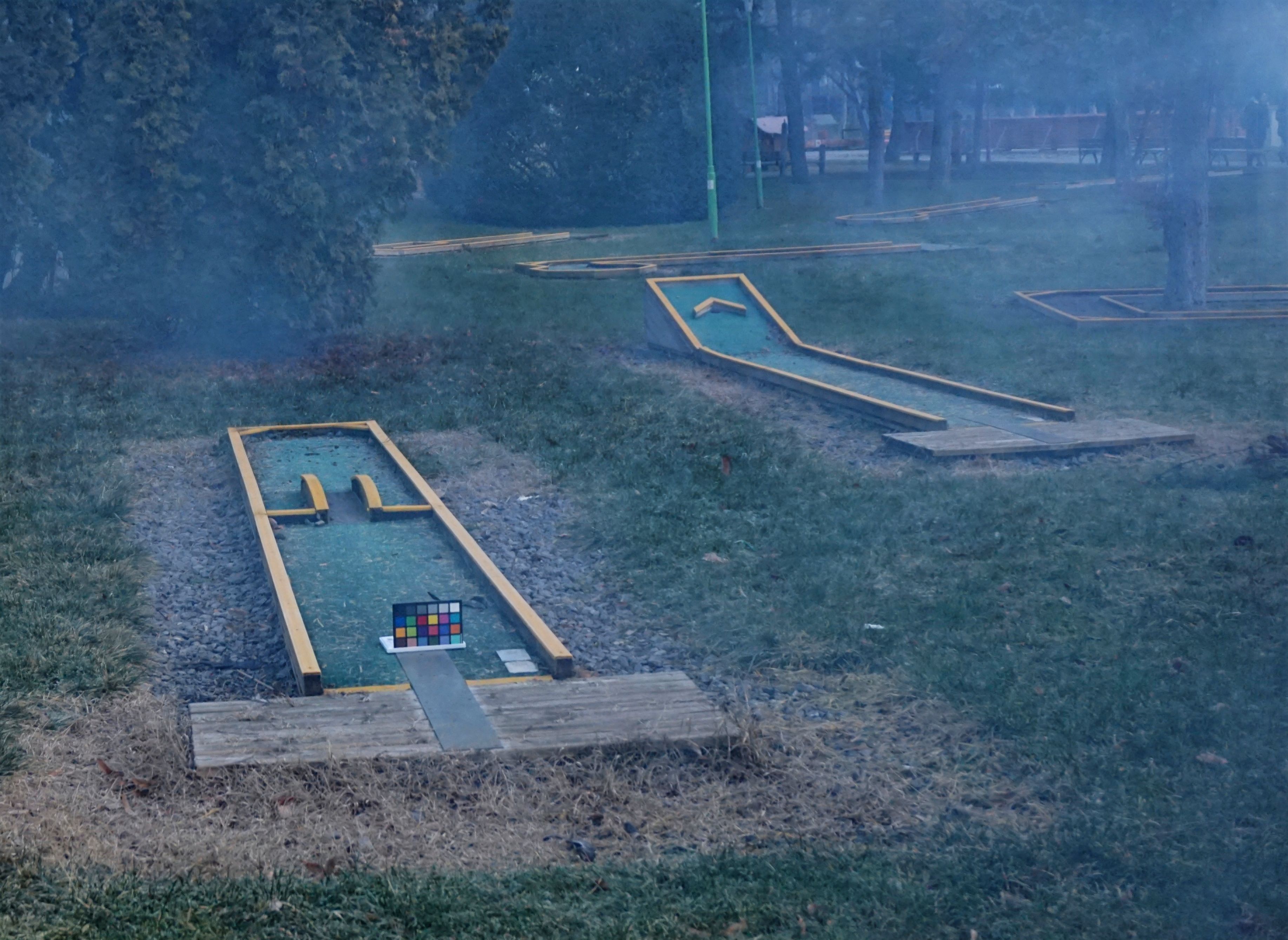}}
 \subfloat{\includegraphics[width=0.24\linewidth]{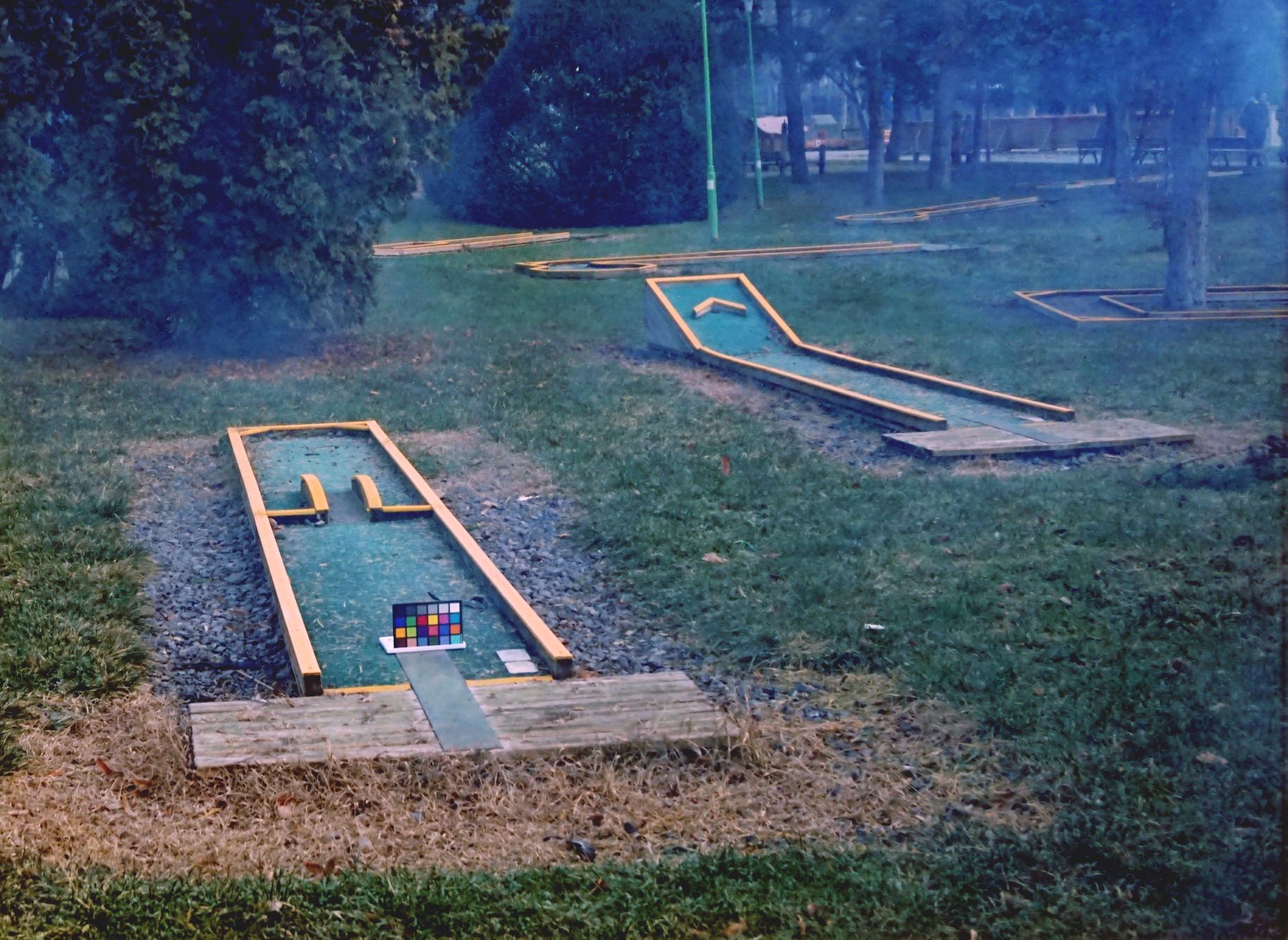}}
 \subfloat{\includegraphics[width=0.24\linewidth]{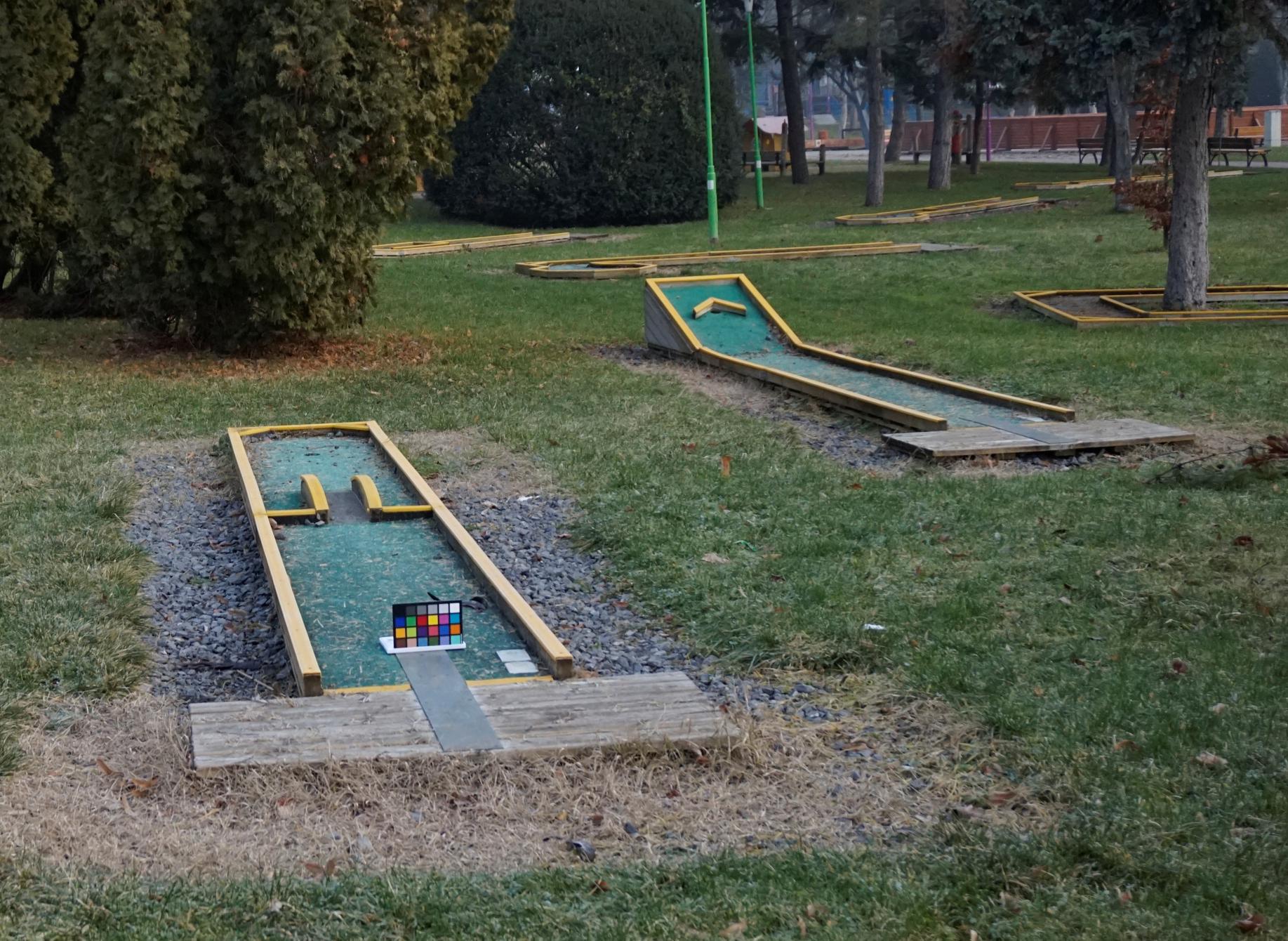}} \\ \vspace{-10pt}
 \setcounter{subfigure}{0}
 \subfloat[Input]{\includegraphics[width=0.24\linewidth]{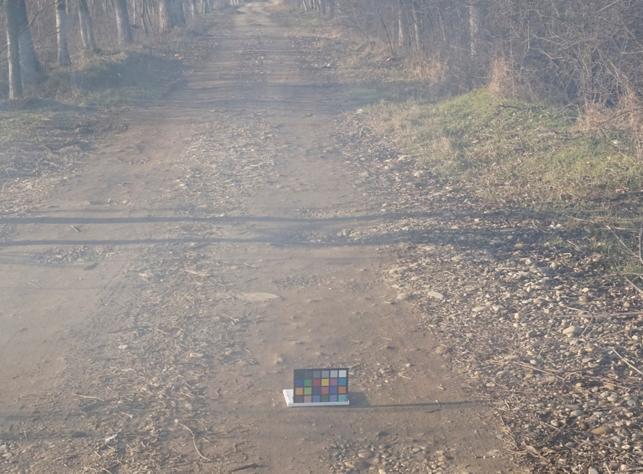}}
 \subfloat[DM2F (RESIDE)]{\includegraphics[width=0.24\linewidth]{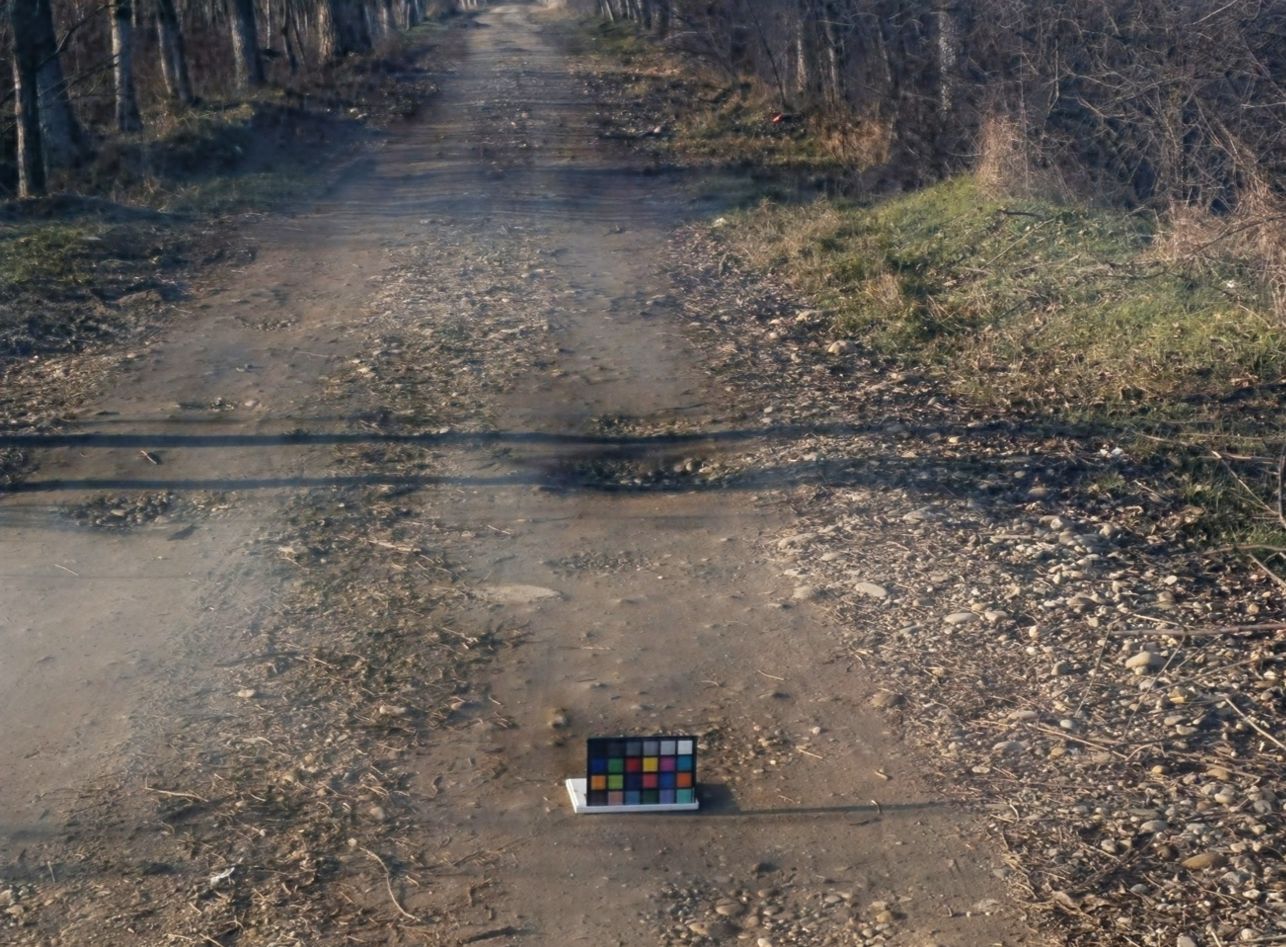}}
 \subfloat[DM2F (LSFD)]{\includegraphics[width=0.24\linewidth]{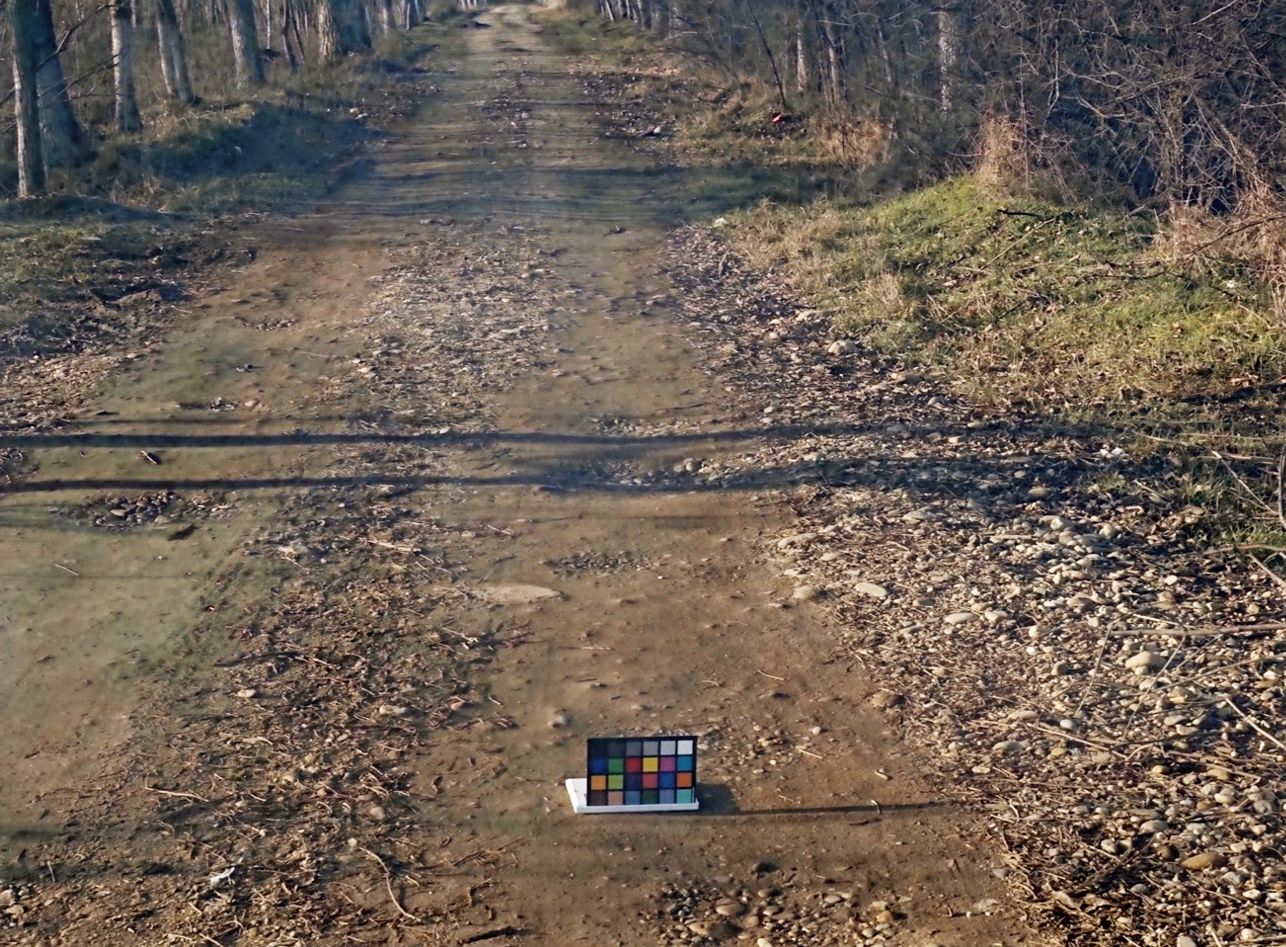}}
 \subfloat[Ground Truth]{\includegraphics[width=0.24\linewidth]{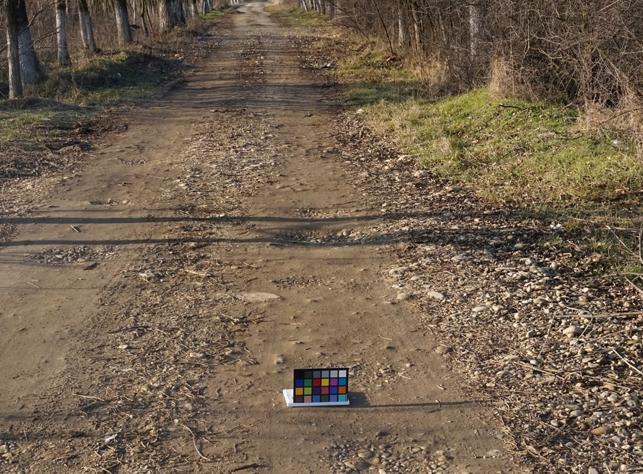}}
 \vspace{-5pt}
 \caption{The figure shows the O-Haze test results of DM2F-Net algorithm trained on RESIDE dataset and the proposed dataset respectively. }
 \label{fig:ohaze}
\vspace{-5pt}
\end{figure*}

\paragraph{Analysis}
To demonstrate the generalizability of the proposed dataset, we have trained recent state of the art methods DF2M-Net~\cite{deng2019deep}, 4K Dehazing~\cite{Zheng_2021_CVPR} both on the proposed LSFD dataset(training set) and RESIDE-OTS~\cite{li2019reside} dataset. And we tested the trained models on another existing high-quality O-Haze~\cite{O_HAZE_2018} dataset.  The quantitative results are shown in Table~\ref{tab:additional_experiment}. One can observe that both of the methods trained on the proposed LSFD dataset obtain better performance evaluated on O-Haze dataset. Visual comparison is also demonstrated in the Fig.~\ref{fig:ohaze}.  More results tested on other existing dehazing datasets and real haze images can be found in the supplementary material. 

\begin{table}[!h]
    \centering
    \small
    \caption{Baseline methods trained on the proposed LSFD dataset and \ruoteng{RESIDE-OTS} dataset,  tested on O-Haze (NTIRE2018) dataset}
    \begin{tabular}{lcccccc}
    \toprule
    Training Data & Method & PSNR & SSIM &  Method & PSNR & SSIM \\ \midrule
    RESIDE-OTS & DF2M-Net~\cite{deng2019deep} & 14.55 & 0.5357 & 4K Dehazing \cite{Zheng_2021_CVPR}  & 15.36 & 0.5792\\
    LFSD (ours) &  DF2M-Net~\cite{deng2019deep} &  16.65 & 0.6380 & 4K Dehazing \cite{Zheng_2021_CVPR} & 17.01 & 0.6836\\
    \bottomrule
    \end{tabular}
    \label{tab:additional_experiment}
    \vspace{-15pt}
\end{table}

\subsection{Experimental Analysis}
In Table~\ref{tab:ourdata_quant}, many recent CNN-based methods such as 4K Dehazing~\cite{Zheng_2021_CVPR}, PSD~\cite{Chen_2021_CVPR} and EPDN~\cite{Qu_2019_CVPR}  methods do not produce convincing results compared with similar state of the art methods \cite{deng2019deep}. Moreover, these methods are even performing worse than traditional dehazing algorithms in the Table. As these methods are pre-trained on other dehazing datasets by respective authors,  most of the performance drop comes from the domain gaps brought about by different datasets. Indicated in their original papers, these methods are usually pre-trained on the RESIDE-OTS dataset, the background of which is quite different from the proposed LSFD dataset. Some of the methods are quite over-fitting to the training data. Although the background images in the RESIDE-SOTS test set are different from those in the training data, then dataset generation process is actually the same, including the depth maps generation and the haze rendering. Therefore, those pre-trained methods are performing good on the RESIDE-SOTS test set, while it does perform badly in other datasets due to clear domain gap. 

\begin{figure*}[t!]
\centering
\vspace{-5pt}
\subfloat{\includegraphics[width=0.33\linewidth]{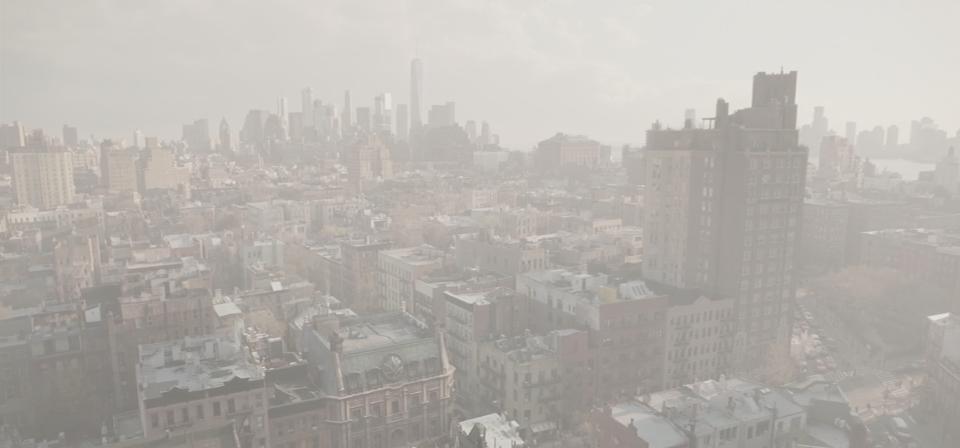}}
\subfloat{\includegraphics[width=0.33\linewidth]{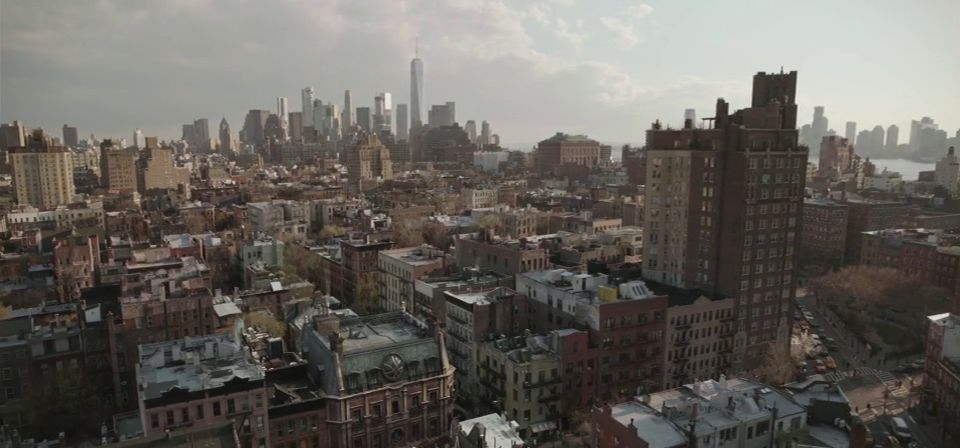}} 
\subfloat{\includegraphics[width=0.33\linewidth]{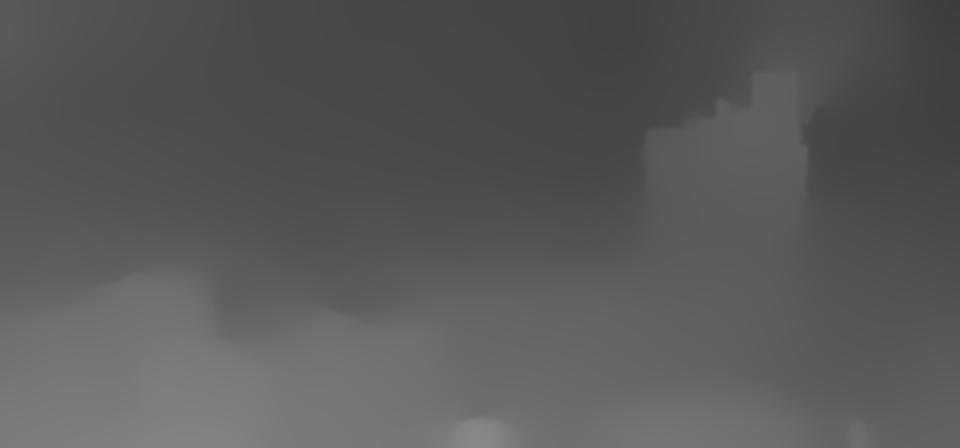}} \\
\subfloat{\includegraphics[width=0.33\linewidth]{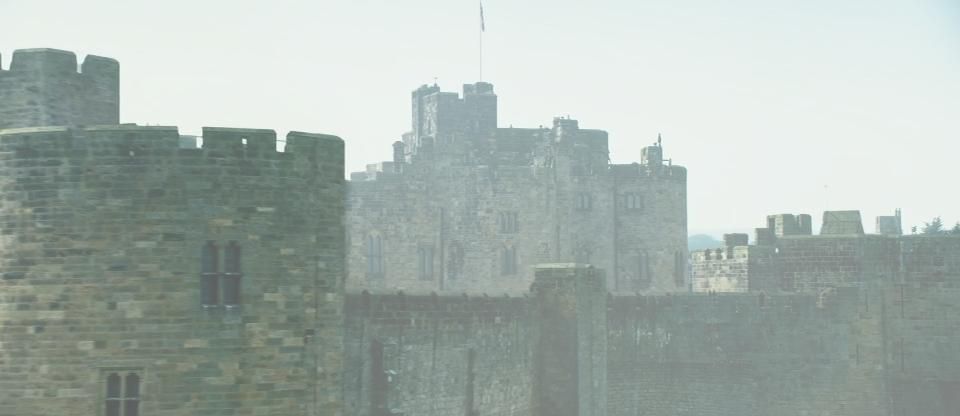}}
\subfloat{\includegraphics[width=0.33\linewidth]{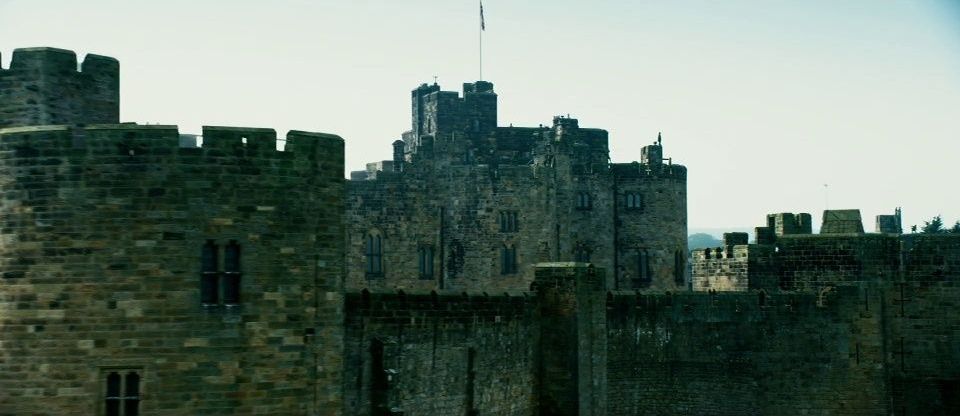}} 
\subfloat{\includegraphics[width=0.33\linewidth]{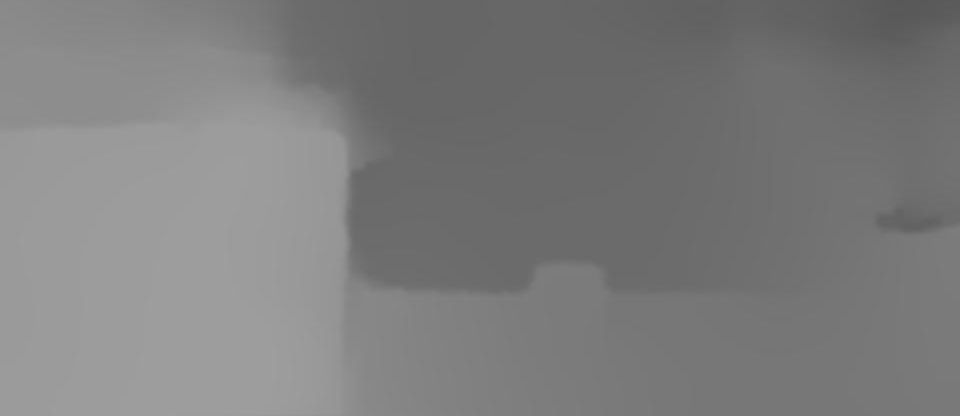}} \\
\caption{Example images of outdoor buildings or skyscrapers in the proposed dataset.}
\label{fig:example_skyscraper}
 \vspace{-15pt}
\end{figure*}

\noindent \textbf{Data distribution comparison with RESIDE} In Fig.~\ref{fig:example_skyscraper}, we have illustrated more city skyscraper examples from the proposed LSFD dataset. These cityscape scenes as well as more diverse modern city views are quite prominent in the proposed dataset. 

\section{Limitation}
The limitation of the proposed dataset can be scrutinized from the data rendering process. As the proposed haze/depth rendering pipeline relies on the optical flow pre-trained model, in theory, the obtained depth maps do not have perfect accuracy compared against physical measurements. However, current technical facilities does not support accurate and dense outdoor depth measurements. In addition, the atmospheric light estimation is also following the widely used MaxRGB illumination algorithm, which could not always handle extreme cases well. 

\section{Conclusion}

We propose a new, large-scale, high-definition and diverse dehazing dataset, which contains real outdoor scenes. We select high quality stereo images of real outdoor scenes and render haze on them using depth from stereo pair. Comparing with previous datasets, our dataset is more realistic due to our high-quality depth. We demonstrate that using this proposed dataset greatly improves the dehazing performance on real scenes. Our dataset is a good complement to existing dehazing datasets and our method provides a practical dehazing solution that is both efficient and effective. We will make the resources available for further dehazing research.

{\small
\newcommand{\cvpr}{IEEE Conference on Computer Vision and Pattern Recognition}
\newcommand{\tpami}{IEEE Transactions on Pattern Analysis and Machine Intelligence}
\newcommand{\nips}{Advances in neural information processing systems}
\newcommand{\threedv}{International Conference on 3D Vision}
\newcommand{\icml}{International Conference on Machine Learning}
\newcommand{\iccv}{IEEE International Conference on Computer Vision}
\newcommand{\eccv}{European Conference on Computer Vision}
\newcommand{\iclr}{International Conference on Learning Representations}
\newcommand{\aistats}{International Conference on Artificial Intelligence and Statistics}
\newcommand{\tip}{IEEE Transactions on Image Processing}
\newcommand{\tog}{ACM Transactions on Graphics}
\newcommand{\ijcv}{International Journal of Computer Vision}
\newcommand{\accv}{Asian Conference on Computer Vision}
\bibliographystyle{unsrt}
\bibliography{egbib}
}


\end{document}